\pdfoutput=1
\documentclass[11pt]{article}

\usepackage[]{acl}

\usepackage{times}
\usepackage{latexsym}

\usepackage{hyperref}
\usepackage{url}
\usepackage{multirow}
\usepackage{booktabs}
\usepackage{enumitem}
\usepackage{xcolor}
\usepackage{amsmath}
\usepackage{flexisym}
\usepackage{bm}
\usepackage{tikz}
\usepackage{subfig}

\usepackage{float}

\usepackage{breqn}

\usepackage[T1]{fontenc}
\usepackage[utf8]{inputenc}

\usepackage{microtype}

\title{Taxonomy Expansion for Named Entity Recognition}

\author{Karthikeyan K$^{1}$\thanks{\textsuperscript{ } Work done during an internship at AWS AI Labs} , {\bf Yogarshi Vyas}$^{2}$\thanks{\textsuperscript{ } Corresponding author}, Jie Ma$^{2}$, Giovanni Paolini$^{2}$, Neha Anna John$^{2}$, \\
\bf Shuai Wang$^{2}$, Yassine Benajiba$^{2}$, Vittorio Castelli$^{2}$, Dan Roth$^{2}$, Miguel Ballesteros$^{2}$ \\
$^{1}$Department of Computer Science, Duke University\\
$^{2}$AWS AI Labs\\
\texttt{karthikeyan.k@duke.edu}, \\
\texttt{\{yogarshi,jieman,paoling,nehajohn,wshui,}\\
\texttt{benajiy,vittorca,drot,ballemig\}@amazon.com}
}

\newcommand{\proposedmethod}{\textsc{PLM}}
\newcommand{\proposedmethodkl}{\textsc{PLM-KL}}
\newcommand{\E}[1]{\mathcal{E}_{#1}}
\newcommand{\D}[1]{\mathcal{D}_{#1}}
\newcommand{\model}[1]{\textit{Model}_{#1}}
\newcommand*\circled[1]{\raisebox{.5pt}{\textcircled{\raisebox{-.9pt} {#1}}}}

\newcommand{\mbcomment}[1]{}
\newcommand{\yvcomment}[1]{}
\newcommand{\kkcomment}[1]{}
\newcommand{\swcomment}[1]{}
\newcommand{\najcomment}[1]{}

\begin{document}

\maketitle

\begin{abstract}

Training a Named Entity Recognition (NER) model often involves fixing a taxonomy of entity types. However, requirements evolve and we might need the NER model to recognize additional entity types. A simple approach is to re-annotate entire dataset with both existing and additional entity types and then train the model on the re-annotated dataset. However, this is an extremely laborious task. To remedy this, we propose a novel approach called \textbf{P}artial \textbf{L}abel \textbf{M}odel (\proposedmethod) that uses only partially annotated datasets. We experiment with $6$ diverse datasets and show that \proposedmethod consistently performs better than most other approaches (0.5--2.5 F1), including in novel settings for taxonomy expansion not considered in prior work. The gap between \proposedmethod~and all other approaches is especially large in settings where there is limited data available for the additional entity types (as much as 11 F1), thus suggesting a more cost effective approaches to taxonomy expansion.
% \yvcomment{Add numbers÷ to abstract.}
% Moreover, we observed that our proposed approach performs significantly better than other baselines in few-shot scenario. Finally, we provide an extensive study analysing the effectiveness of our approach in various scenarios.
\end{abstract}

\section{Introduction}

\begin{figure}[!t]
    \centering
    \includegraphics[width=0.4\textwidth,trim={9.8cm 5cm 13.3cm 3cm},clip]{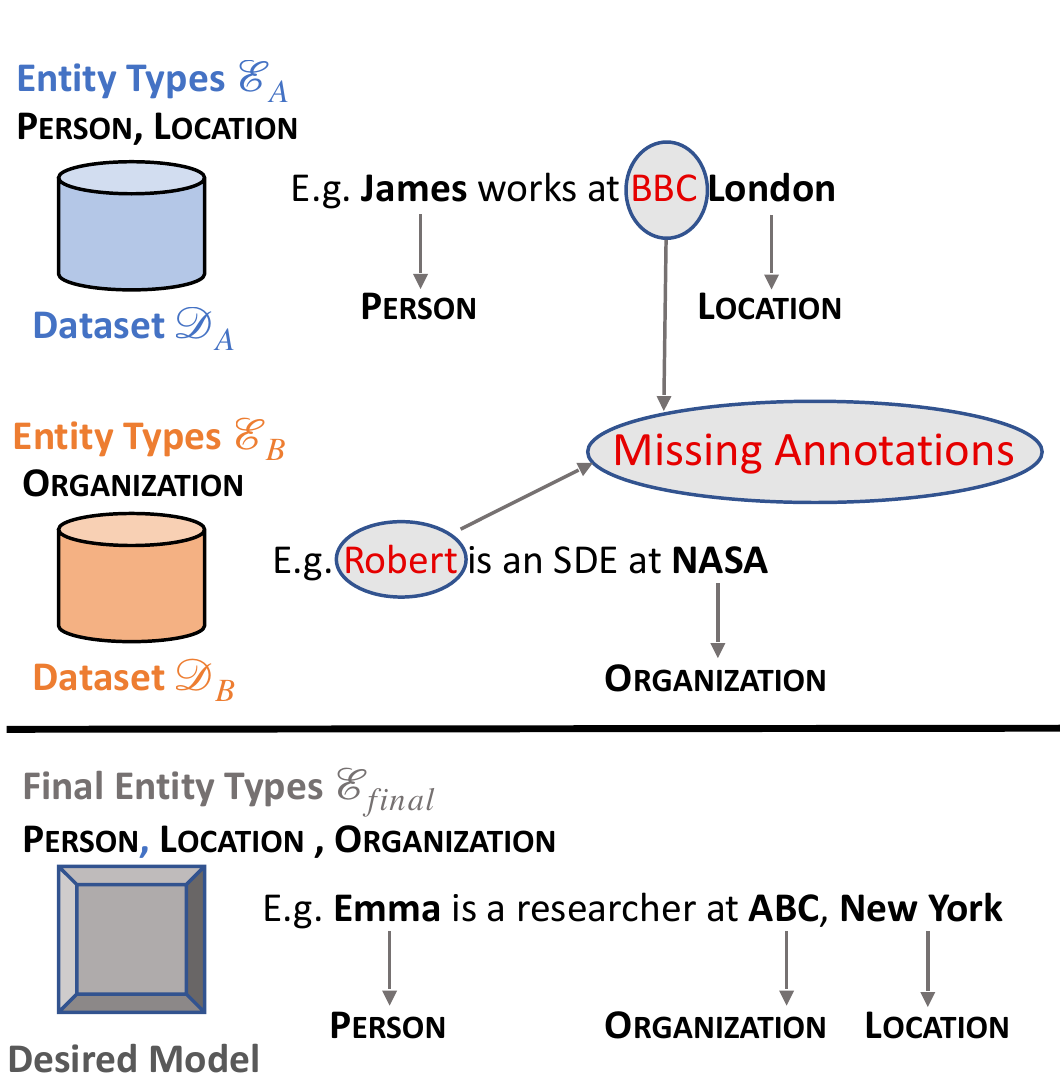}
    \caption{\textbf{Taxonomy Expansion for NER:} Given Dataset $\D{A}$ annotated only with entity types $\E{A}$ and $\D{B}$ annotated only with $\E{B}$, the task is to train a model that recognize entities from both $\E{A}$ and $\E{B}$.}
    %\swcomment{minor: add `s' to Final Entity Type`s' in the figure}}
    \label{fig:example}
\end{figure}

% High-level problem
Training a Named Entity Recognition (NER) model typically involves presupposing a fixed taxonomy, that is, the entity types that are being recognized are known and fixed. However, NER models often need to extract new entity types of interest appearing in new text, thus requiring an expansion of the taxonomy under consideration. Consider an example (Figure~\ref{fig:example}), where an NER model was initially trained to recognize \textsc{Person} and \textsc{Location} types using data annotated only for these two types. However, it later needs to identify \textsc{Organization} entities in addition to the two initial entity types. 

There are several simple ways to solve this problem, yet they are cumbersome in different ways. One possible solution is to re-annotate all data that was only annotated for the initial entity types (\textsc{Person}, \textsc{Location}) with the new entity types (\textsc{Organization}), and then train a single model to recognize all the entity types on this data. If such re-annotation and re-training needs to happen iteratively every time new entities are introduced, then this is an expensive process, and re-annotation requires knowledge of all entity types both old and new.
A secondary issue with this approach is that the original training data was chosen only to contain examples of the initial entity types. Thus, it may not even contain enough examples of the new entity types and it may be simpler to obtain data that is rich in the new entities types for annotation (\textsc{Organization}).  
%\swcomment{Here I think we should probably also explain/motivate why it's not simple/easy to have old data types in new dataset too, i.e., why Robert/PERSON is not labeled? as shown in the example} 
Given these two disjoint datasets with two separate sets of annotations, we can train two different models to recognize the two sets of entity types --- yet this requires heuristics to combine the predictions of the two models, especially if the old and new entity types have some shared semantics (e.g. some of the new entity types are subtypes of the existing entity types). 
 %Another approach might be to train two separate models, one for the original entity types and another for the new entity type.   \mbcomment{or to train 2 models, right? we should explain what the issue will be if we do that. Btw, maybe we should have that baseline.} 

% What has prior work looked at in this space? + Our contribution of definition 
This problem of Taxonomy Expansion for NER (henceforth, TE-NER) has been studied in several different settings in prior work.
% \citet{Monaikul_Castellucci_Filice_Rokhlenko_2021} and \citet{xia-etal-2022-learn} study TE-NER under the assumption that they only have access to a model trained on the original entity types as well as annotations for the new entity types on a different dataset.
However, most prior work (\S\ref{sec:relatedwork}) assumes that the old and the new entity types are completely disjoint and mutually exclusive. In contrast, we define a general version of the TE-NER problem (\S\ref{sec:defintion}) where the original training data can be used, and old and new entity types can be related in different ways. Our general setup allows for many practical scenarios encountered during taxonomy expansion such as new entity types being subtypes of existing entity types, or partially overlapping in definition. 

% Model contribution 
We then propose a novel solution to TE-NER motivated by the fact that the available training data are only partially annotated (\S\ref{sec:proposed_method}). As shown in Figure~\ref{fig:example}, \underline{\textbf{BBC}} has not been annotated 
%\mbcomment{not true because dataset A never had that in its taxonomy, we need to improve the writing of this section} 
as \textsc{Organization} in  the initial dataset $\D{A}$, since \textsc{Organization} was not in the label taxonomy of  $A$. Similarly, \underline{\textbf{Robert}} is not annotated as \textsc{Person} in the new dataset $\D{B}$. Given such partial annotations, we treat the true, unobserved label as a latent variable and derive our approach called \textbf{P}artial \textbf{L}abel \textbf{M}odel (\proposedmethod). 
%\swcomment{Not very clear why latent variable is needed? `Given such partial annotations' seems not a direct explanation for introducing a latent variable? Add something short about it may help the reasoning.}  \yvcomment{added}
Intuitively, \proposedmethod~uses a model trained on $\D{B}$ to annotate \underline{\textbf{BBC}} with a probability distribution over entity types $B$.
Similarly, it uses a model trained on $\D{A}$ to obtain predictions for \underline{\textbf{Robert}}.
Finally, the desired model over all entity types is trained using both the observed annotation and the soft labels obtained via the predictions of the intermediate models. 
%In contrast to \citet{Monaikul_Castellucci_Filice_Rokhlenko_2021} and \citet{xia-etal-2022-learn}, we use a model trained on dataset $B$ to (softly) annotate examples in dataset $A$ and train our desired model on combined dataset $A$ and $B$ and not just dataset $B$. 
Minimizing the KL divergence loss (as used by \citet{Monaikul_Castellucci_Filice_Rokhlenko_2021} and \citet{xia-etal-2022-learn}) corresponds to maximizing a lower bound on the likelihood of our formulation, while the proposed \proposedmethod~loss corresponds to directly maximizing the likelihood.

%\yvcomment{let's add a sentence here on how the model works intuitively}.
%we also propose an intuitive baseline called \emph{Cross Annotation}. Despite being simple, our experiments show that \emph{Cross Annotation} is a very strong baseline that performs competitively with our proposed method. 

%% Experimental findings
Experiments with six English datasets show that \proposedmethod~outperforms all baselines except one (cross annotation) by 0.5--2.5 F1 (\S\ref{sec:experiments}).
However, this competitive baseline, that involves simply \textit{cross-annotating} the datasets using weak labels from a model trained on the other dataset, is less accurate in few-shot settings compared to \proposedmethod. When only 200 examples are provided for each new entity type, \proposedmethod~is 11 F1 better than cross-annotation. We further test the robustness of these approaches across several data distributions and find that they perform robustly across the board (\S\ref{sec:analysis}).
%\kkcomment{than other baselines. (??)}

%We observe that both \proposedmethod and cross annotation are (1) equally effective when entity type $A$ and $B$ contains $10$ and $10$ or $19$ and $1$ entity types, (2) better at recognizing entity types $A$ better than the model trained only on entity types $A$, Furthermore, we also observe that \proposedmethod is much better than cross annotation in limited data settings\mbcomment{unclear what "limited data settings" means}. Finally, we propose a validation strategy\mbcomment{explain that this is the validation set used for stopping criteria} that uses two partially annotated validation data (just like training datasets) and show that even with a partially annotated validation dataset, proposed methods perform as good as with a fully annotated validation data. 

% In summary,
% \begin{enumerate}[nosep]
%     \item We define \emph{``Taxonomy Expansion for NER"} (TE-NER); different from prior work, our definition assumes access to the original entity types and does not assume the novel entity types to be mutually exclusive from the original entity types.  
%     \item We propose a novel approach called \proposedmethod~to solve the TE-NER problem. Experiments with 6 different datasets show that \proposedmethod~performs consistently better than all but one baselines, and is significantly more accurate when less training data is available.
%     \item We analyse the robustness of our approach in various scenarios \yvcomment{TBD based on whether we include the last section}.
% \end{enumerate}
\section{Related Work}
\label{sec:relatedwork}

While this work is closely related to knowledge distillation~\citep{hinton2015distilling,shmelkov2017incremental} and continual learning (CL) based methods for NER \citep{xia-etal-2022-learn,Monaikul_Castellucci_Filice_Rokhlenko_2021,chen2019transfer,de2019continual}, our setup differs in two aspects from previous tasks. 

First, we assume access to both initial ($\D{A}$) and new datasets ($\D{B}$), while typical CL based methods assume access only to the new data and to a model trained on the initial data. \citet{xia-etal-2022-learn} assumes access to a generative model trained on the initial dataset and trains the final model in a two-stage process. In contrast, we systematically derive \proposedmethod~using a latent variable formulation and train our final model in a single stage. 

Second, we do not constrain the new entity types to be disjoint from old entity types. For example, if \textsc{Person} is among the initial entity types, the new entity types can include \textsc{Actor}, which is a sub-type of \textsc{Person}; our desired final model should predict a mention as (\textsc{Actor}, \textsc{Person}) if the mention is an actor, and just \textsc{Person} if the mention is a person but not an actor. This setup is related to prior work on hierarchical classification~\citep{silla2011survey,arabie1996hierarchical,meng2019weakly} or fine-grained NER~\citep{ling2012fine,choi2018ultra,mai-etal-2018-empirical,ringland2019nne}.
However, the objective of such work is to use the hierarchical structure to train a better fine-grained NER model; they do not deal with taxonomy expansion. To the best of our knowledge, our work is the first to handle non-disjoint entity types in the context of taxonomy expansion. 

Another class of related approaches is based on positive-unlabelled (PU) learning~\citep{bekker2020learning,grave2014weakly,peng-etal-2019-distantly} or learning from partial/noisy training data~\citep{mayhew-etal-2019-named} ---  if we combine the initial ($\D{A}$) and new datasets ($\D{B}$), then the combined training data can be viewed as a positive-unlabelled or noisy dataset. However, unlike PU learning, the annotations in our setup are not randomly missing; instead, annotations of entity types $A$ and $B$ are missing in datasets $\D{B}$ and $\D{A}$ respectively. 
\section{Problem Definition and Notation}  \label{sec:defintion}

Given datasets $\D{A}$ and $\D{B}$ annotated with entity types $\E{A}=\{ e_A^1, e_A^2, ... e_A^m \}$ and $\E{B}=\{ e_B^1, e_B^2, ... e_B^n \}$ respectively, the goal of TE-NER is to learn a model to recognize entities from both $\E{A}$ and $\E{B}$. %\swcomment{explain n and m? number of entity types for them respectively? }
For our initial definition and solution, we assume that all entity types are distinct and mutually exclusive, i.e., a token does not belong to more than one entity type. While this is in line with prior work, it is also a shortcoming as entity types and definitions can often overlap in real world settings as well as academic datasets. For instance, in the FewNERD dataset~\cite{ding2021few}, an entity can be both a \textsc{Person} as well as a \textsc{Politician}. 
%Similarly, \textsc{Locations} in the Few-NERD dataset are not identical to \textsc{Locations} in Ontonotes, as the latter does not include GPEs. 
%\mbcomment{we need to explain why this is an important limitation, and why it is still relevant to study this}\yvcomment{Done.} 
To handle such cases, we define a more general version of the problem and solutions in \S\ref{sec:revisit}.
%$\E}_{final} = $ 
%. We assume  all the entity types are distinct and relationship ($\R}$), between any pair of entity types are disjoint, i.e a mention (in a given context), does not belong to more than one entity type. For example, \textsc{Person} and \textsc{Organization} are disjoint, whereas \textsc{Person} and \textsc{Actor} are not. Refer ~\ref{fig:example} for examples. 

Prior work on TE-NER is based on continual learning where the goal is to adapt the NER model continuously to new entity types~\citep{Monaikul_Castellucci_Filice_Rokhlenko_2021,xia-etal-2022-learn}. Such work assumes access to only the models trained on the original datasets ($\model{A}$ trained on $\D{A}$, $\model{B}$ trained on $\D{B}$) and not the datasets themselves. 
% However, the problem of TE-NER is also crucial in settings where two parties have access to different sets of annotations and entity types (e.g., a commercial NER service offers a set of entity types out of the box, but a specific customer wants to extend it to additional entity types). 
However, in many scenarios, both the original dataset and the new dataset are available but are annotated using different sets of entity types due to an evolving taxonomy or come from different sources.
%there is no strong justification to assume such settings. \najcomment{Is this true? I suppose there could be scenarios where you do not have access to the data, but just the models?}\yvcomment{Rephrased} In fact, . 
Hence, our definition assumes full access to both $\D{A}$ and $\D{B}$.
%\swcomment{following the commercial NER use case, maybe we should briefly indicate the extension happens on the commercial NER end (like a custom NER service) since the customers themselves may not have full access to $\D{A}$ (internal data for training the commercial NER model)?}\yvcomment{I'm not sure it is relevant which side the extension happens on, since the problem is symmetric.}
Also, unlike few-shot and transfer-learning setups~\citep{Phang2018SentenceEO,ma-etal-2022-label}, our goal is to train a model ($\textit{Model}_{final}$) that does well on both $\E{A}$ and $\E{B}$ (irrespective of the size of $\D{B}$; however, we return to the question of size in \S\ref{sec:exp_rq2}).

%\yvcomment{I think we need some sentences here on why we choose to frame it in this way, and use that to contextualize with respect to prior work. }

\paragraph{Why is Taxonomy Expansion challenging?} The central challenge with TE-NER is partial annotations --- if a mention in $\D{A}$ belongs to an entity in $\E{B}$, it will not be annotated (e.g. in Fig~\ref{fig:example}, \underline{\textbf{BBC}} is not annotated as \textsc{Organization}). Similarly, $\D{B}$ is also partially annotated. Such partial annotation misleads model training, and prior work attempts to mitigate this issue on a single dataset \cite{mayhew-etal-2019-named,jie-etal-2019-better}. We focus on this problem in the context of TE-NER.

\section{Methods for TE-NER}

\begin{figure}[!tbp]
    \centering
    \includegraphics[width=0.37\textwidth,trim={9.8cm 5cm 9.5cm 4cm},clip]{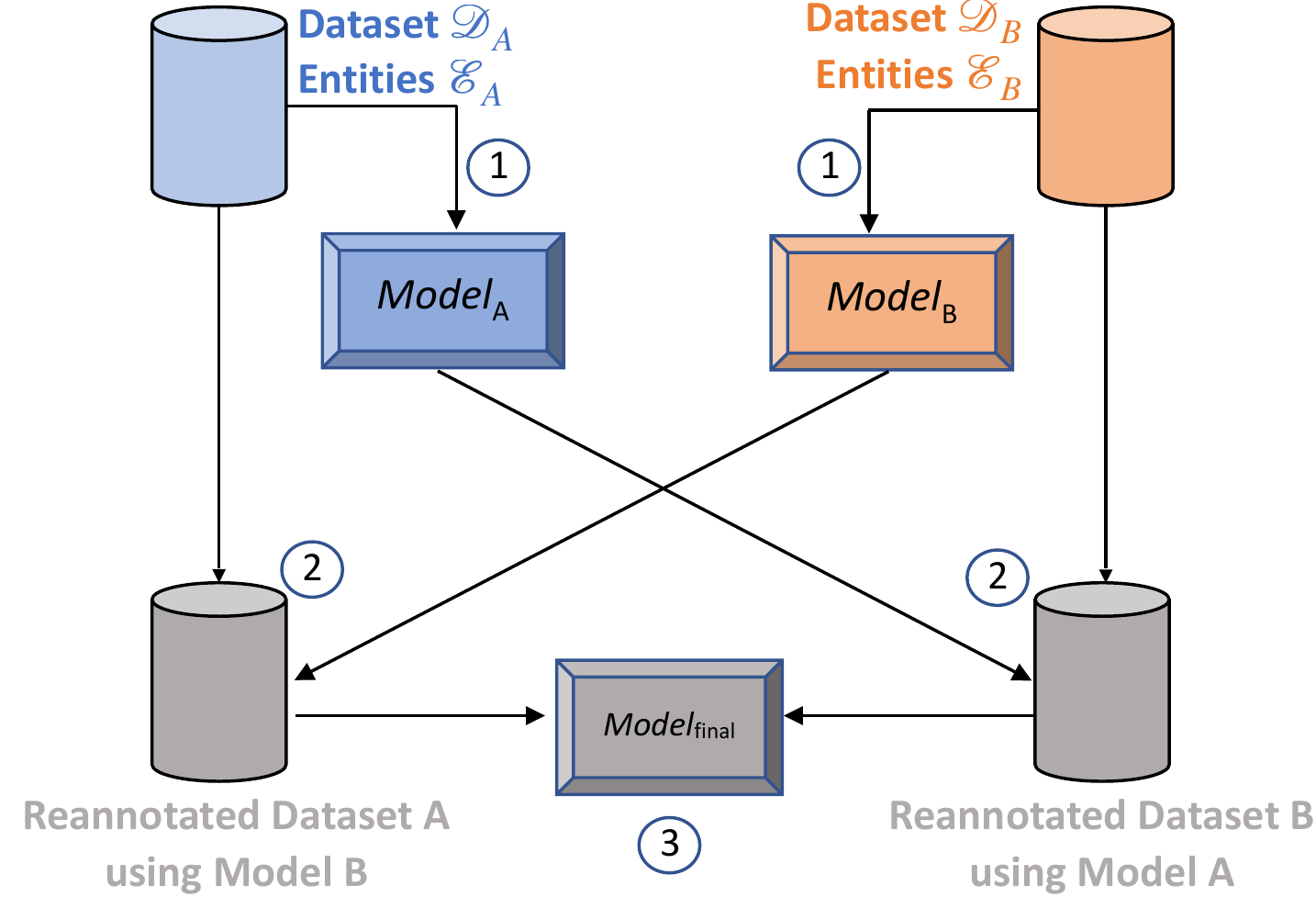}
    \caption{\proposedmethod~consists of three steps: \circled{1} Train $\model{A}$/$\model{B}$ to recognize entities in $\E{A}$/$\E{B}$ using dataset $\D{A}$/$\D{B}$ \circled{2}Use $\model{A}$ to annotate $\D{B}$ with distribution over entity types $\E{A}$ (repeat with $\model{B}$ and $\D{A}$ \circled{3} Use these annotations with the \proposedmethod~loss to train the final model. Cross-annotation is similar except in \circled{2}, we annotate the dataset with the hard predictions of the models instead of a distribution, and use a cross-entropy loss in \circled{3}. \proposedmethodkl~replaces the \proposedmethod~loss in \circled{3} with a KL-divergence term.}
    % \textbf{Cross Annotation, \proposedmethod, \proposedmethodkl and CL Method:} Illustration of Cross Annotation and \proposedmethod.
    % CL method is, just step 1, 3.B and 4 of \proposedmethod-KL (final model is trained only on Dataset B with probabilities from Model A).\kkcomment{Explain steps in caption ? Remove right half?} }\najcomment {Didn't understand the description of the figure here. I suppose this will be updated, based on Karthik's comment}
    \label{fig:methods}
\end{figure}

We first discuss our methods for the scenario where all entity types are disjoint; then, in \S\ref{sec:re_proposed_methods}, we discuss modifications for a more general definition of TE-NER. We use the BIO scheme \cite{ramshaw-marcus-1995-text} for this work. All the approaches below are motivated by a simple observation: in both $\D{A}$ and $\D{B}$, the observed labels are not always the true labels. Specifically, if a word is annotated as O (used to indicate tokens that are not part of entity mentions), we do not know its true label.  Figure~\ref{fig:methods} provides an overview of all approaches.

\subsection{Cross Annotation (X-Ann.)}
\label{subsec:crossann}

Before discussing our solution, we suggest a naive solution to TE-NER. 
As discussed before, the main challenge with TE-NER is partial annotations. If the annotations are correct and exhaustive, i.e.,  $\D{A}$  was annotated for entity types $\E{B}$ (and $\D{B}$ for $\E{A}$), then we could simply combine these datasets and train an NER model on the combined dataset. Cross-annotation is motivated by this observation --- instead of expecting the data to be fully labeled, it uses model outputs to provide the missing labels. 

Under our setup, if a token in $\D{A}$ is annotated as O, it can still belong to either one of the entities in $\E{B}$ (e.g. \textbf{BBC} in Figure~\ref{fig:example} can truly be O or it can be an \textsc{Organization}). Therefore, in $\D{A}$, if a word is annotated as O, we replace it with the prediction of $\model{B}$. Similarly, in $\D{B}$, if a word is annotated as O, we replace it with the prediction from $\model{A}$. We combine these re-annotated versions of $\D{A}$ and $\D{B}$ to train the final model.

\subsection{Partial Label Model (PLM) } \label{sec:proposed_method}

Cross annotation uses the hard labels obtained from the predictions of $\model{A}$ and $\model{B}$ to re-annotate $\D{A}$ and $\D{B}$. We present an alternate approach that extends cross-annotation to use \textit{soft} labels (i.e., the entire output distribution).
%and negative log dot product or KL divergence loss instead of CE loss. 
%Fig~\ref{fig:methods} \swcomment{Minor: we sometimes use Figure and sometimes use Fig. Choose one and make them consistent?} provides an illustration of this. \yvcomment{Need to work on Figure to make this true.}
% \yvcomment{Need to make connections here with student-teacher learning.}

%The cross-annotation approach proposed above has obvious pitfalls -- \yvcomment{list pitfalls of the approach}

% \yvcomment{I think this latter sentence is something that generally connects both cross-annotation and PLM. Let's move this obsercation to the first paragraph, of this section, present cross-annotation as a naive way of solving this problem, and then our method as an improvement on the naive way.}

As in cross annotation, we start with the observation that the observed label for a token is not necessarily the true label, and conversely, the true label is unobserved. Thus, we treat the the true label as a latent variable. 
%\swcomment{Add some motivation/explanation why we introduce/use a latent variable for it? Behind intuition?} \yvcomment{done}
First, to simplify the discussion, we assume that the distribution of this latent variable is known and then solve for the optimal parameters of the model. Later, we relax this assumption and discuss how to approximate the distribution of this latent variable. 

Let us denote our desired final model by $\model{\textit{final}}$, parameterized by $\theta$ as $f(s|\theta)$. The output of $f(\cdot)$ is a probability distribution over the entity labels $\E{\textit{final}} = \E{A} \cup \E{B}$, for each token in the input $s$ and the output for $i^{th}$ token is denoted by $f_i(\cdot)$. We use a BERT-based sequence tagging model for $f(\cdot)$ (\S\ref{sec:dataset_creation}). 

First, let us calculate the likelihood for a single example $s\in\D{A}\cup\D{B}$ consisting of $n$ tokens, $s = [w_1, w_2, \dots, w_n]$, and its corresponding observed labels $y = [y_1, y_2, \dots, y_n]$. We denote the corresponding (latent) true labels as $Z =  [Z_1, Z_2, \dots, Z_n]$ and predicted labels as $Y =  [Y_1, Y_2, \dots, Y_n]$ with $Y_i \sim f_i(s| \theta)$ and $Z_i \sim g_i(s)$. Here $g()$ is an oracle function that gives us the distribution of true labels. The likelihood of the predictions matching the true underlying label, $P(Y= Z|\theta)$, is then calculated as
\begin{align*}
    P(Y = Z | \theta_{final} ) = \prod_{i=1}^{n} P(Y_i = Z_i).
\end{align*}
Further, $P(Y_i = Z_i)$ can be decomposed as
\begin{align*}
    P(Y_i = Z_i) &= \sum_{e \in \mathcal{E}_{final} } f_i(s|\theta)[e]\times g_i(s)[e] \\
    &= 	\Big \langle \ f_i(s|\theta),\ g_i(s)\ \Big\rangle.
\end{align*}
Finally, the negative log-likelihood yields the loss:
\begin{align}
    Loss(s|\theta) &= \sum_{i=1}^n -\log{ \Big \langle \ f_i(s|\theta),\ g_i(s)\ \Big\rangle } \label{eq:loss_dotproduct}.
\end{align}
\paragraph{Approximating $\bm{g(\cdot)}$} Equation~\ref{eq:loss_dotproduct} assumes access to an oracle function $g(\cdot)$ that gives us the distribution of the true labels. In practice, this is exactly the information that we do not have access to. Thus, we approximate $g$ using the predictions of a model, based on two key observations.

First, if a token $w_i \in s$  (for $s \in \D{A}$) is annotated as an entity $e$, we know that the annotated label is the true label. In this case, $g_i(s)[e] = 1$, and $g_i(s)[e'] = 0~\forall e' \neq e$.
%and hence the distribution of true label (i.e. $1$ for the annotated label and $0$ for rest all). 
Second, if $w_i$ is not annotated as an entity (i.e., is assigned an O label), then we know that it does not belong to any entity in $\E{A}$. Therefore $g_i(s)[e] = 0 \ \forall e \in \E{A}$. However, $w_i$ could potentially be an entity $\in \E{B}$. So, we just need the probability distribution over $\E{B}\cup \{O\}$,\footnote{More precisely, a token is not annotated with an entity type (e.g. \textsc{Person}), but a combination of B/I tag and the type (e.g. \textsc{B-Person}). We drop the B/I tags for simplicity, but the approach works identically regardless of the tags.}
%\swcomment{strictly speaking $\E{B}$ is Entity type Set e.g.,  PERSON, but not PERSON-B (since here the label is for each token $w_i$)? But this should be fine. I indicated this in case we want to be more/strictly accurate. I think one simple solution is to add a footnote, about the token-level label is PERSON-B but for illustration/simplicity we use PERSON.}\yvcomment{Good point, done thanks}
which we can directly estimate by using $\model{B}$.\footnote{$\model{B}$ gives us the probability of not belonging to any of the $\E{B}$, but since we already know that the token does not belong to $\E{A}$, the probability of not belonging to $\E{B}$ is equal to the probability of not belonging to $\E{A}$ and $\E{B}$.} Analogously, for $s \in \D{B}$, we can use $\model{B}$ to estimate the distribution over $\E{A}$. With these approximations, Equation \ref{eq:loss_dotproduct} can be split into two terms corresponding to the two cases above. For the first case, the loss is simply a cross-entropy loss against the one-hot vector obtained from $g_i(s)$. This gives us the proposed loss function of \proposedmethod. 
\begin{align}
    Loss(s|\theta) &=  \sum_{i:y_i = O} CE \Big(f_i(s|\theta),\ y_i \Big) \nonumber \\
    &+ \sum_{i:y_i \neq O} -\log{ \Big \langle \ f_i(s|\theta),\ g_i(s)\ \Big\rangle }. \label{eq:keyloss}
\end{align}
\paragraph{\proposedmethodkl:} The loss term in Equation~\ref{eq:keyloss} is similar to the knowledge distillation loss, where the second term is replaced by a KL-divergence term:
%If we replace the second term in Equation \ref{eq:keyloss} with a KL-divergence term 
%\swcomment{may want to indicate any behind reason/motivation we propose to use KL as another variant?}\yvcomment{done}, 
%we arrive at a loss function identical to the one used in previous work~\cite{} in a continuous learning framework. 
\begin{align}
    Loss(s|\theta) & =  \sum_{i:y_i = O} CE \Big(f_i(s|\theta),\ y_i \Big) \nonumber \\
    &+ \sum_{i:y_i \neq O}  KL\Big ( g_i(s) \ || \ f_i(s|\theta) \Big). \label{eq:lowerbound}
\end{align}
This loss term has been used in prior work on continuous learning in NER~\cite{Monaikul_Castellucci_Filice_Rokhlenko_2021} and it simulates a student-teacher framework with $g$ as the teacher, and $f$ as the student. We prove, using a simple application of Jensen's inequality, that the loss in Equation~\ref{eq:lowerbound} is an upper bound of the loss in Equation~\ref{eq:keyloss} (Appendix~\ref{appdx:proposed_method}). In \S\ref{sec:experiments}, we experiment with both the exact loss function of \proposedmethod~as well as the upper-bound of \proposedmethodkl.
\section{Revisiting Taxonomy Expansion}
\label{sec:revisit}

\subsection{Definition} \label{sec:re_defintion}
\begin{figure}[t]
\centering
\subfloat[Illustration of redefined output space]{
\label{fig:relations}
\includegraphics[scale=0.4,trim={3cm 5cm 11cm 0.5cm},clip]{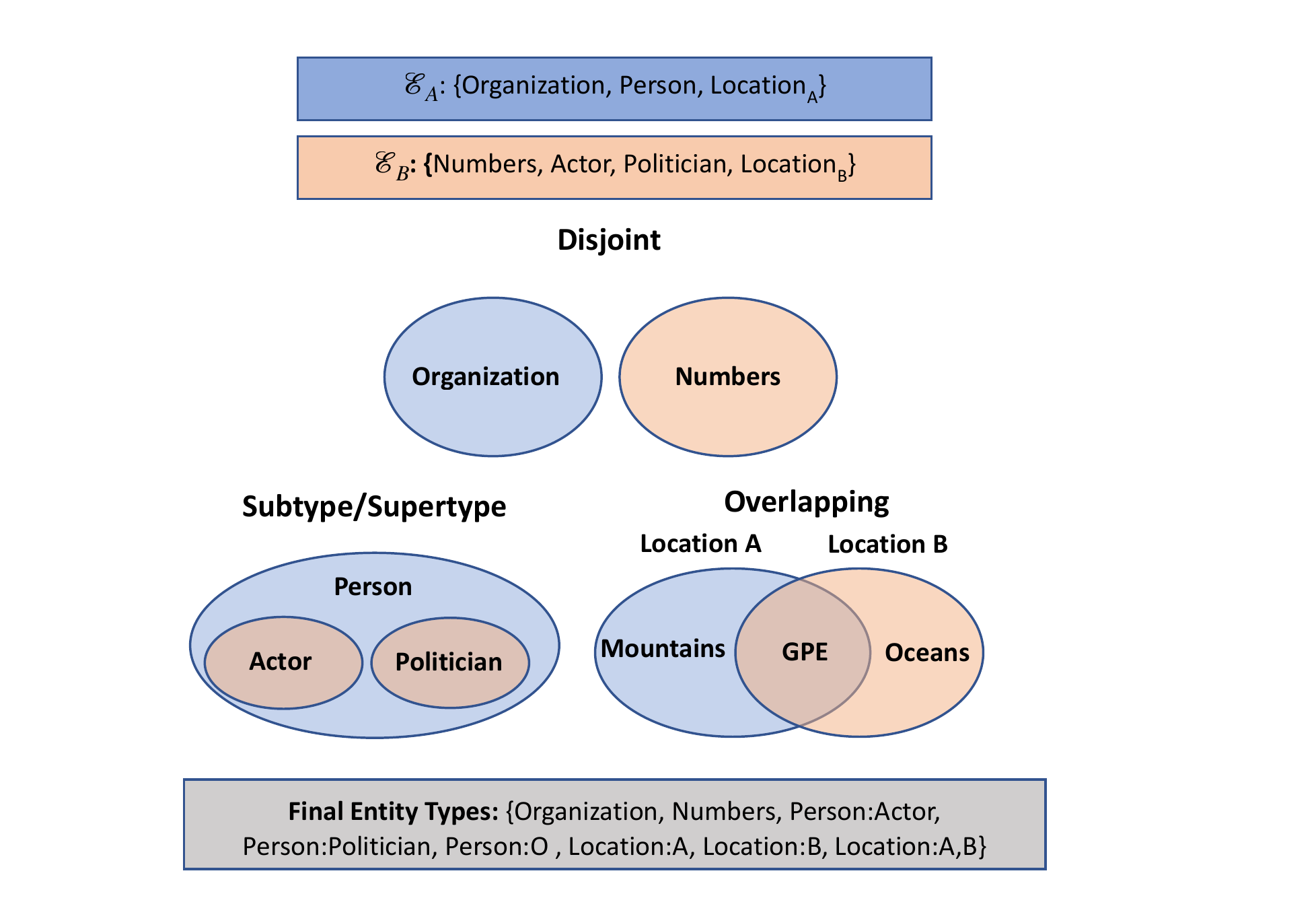}}
\vspace{0.1cm}
\subfloat[Illustration of allowed entity types]{
\label{fig:allowed}
\includegraphics[scale=0.4,trim={0cm 11cm 13cm 0.7cm},clip]{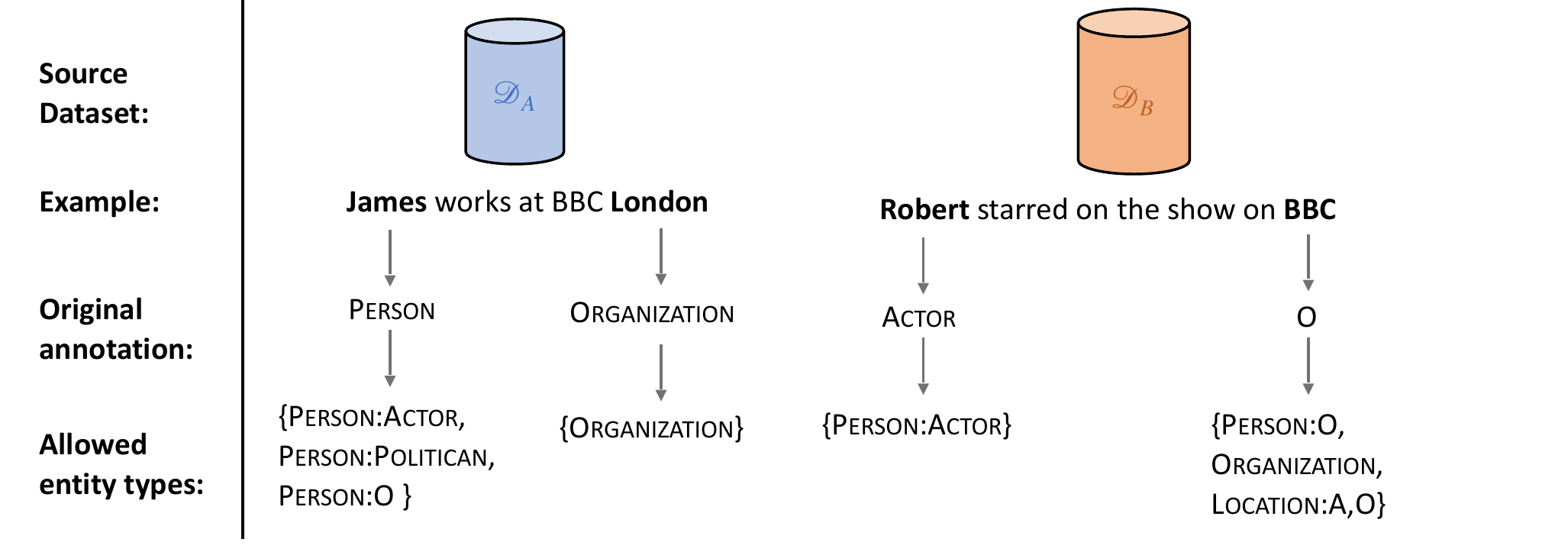}}
\caption{For the non-disjoint case, the output space of the combined entity types is not simply a union of the two original entity types. \ref{fig:relations} illustrates how the final entity types look like given the original types and the relations between them. \ref{fig:allowed} shows the allowed entity types in the redefined output space for each token in the dataset, given their original annotations.}
\end{figure}

% \begin{figure*}
%     \centering
%     \includegraphics[width=\textwidth]{figures/relation_allowed_labelmaps.png}
%     \caption{\textbf{Relationship, Allowed Entity Types and Label Maps}\kkcomment{Better ideas for figure? and explain in caption} \yvcomment{Need to work on figure to make it more clear}}
%     \label{fig:relation_allowed_labelmaps}
% \end{figure*}

In \S\ref{sec:defintion}, following prior work, we assumed that all the entity types are distinct and disjoint. We now extend our definition of the problem as well as methods to a more general version which allows for semantic overlap in the entity types. Specifically, we assume that along with $\D{A}, \D{B}, \E{A}$, and $\E{B}$, we also know the relationship $\mathcal{R}$, between the entity types ($\E{} = \E{A} \cup \E{B}$), where
$ \mathcal{R}: \E{} \times \E{} \rightarrow$  \{\textsc{Disjoint}, \textsc{Subtype}, \textsc{Supertype}, \textsc{Overlapping}\}.
The \textsc{Subtype}/\textsc{Supertype} relations allow for an entity in $\E{B}$ (e.g. Politician) to be a subtype of an entity in $\E{A}$ (e.g. Person), and vice versa. The \textsc{Overlapping} relation allows for partial overlaps in the definitions of types (e.g. if both $\E{A}$ and $\E{B}$ have a \textsc{Location} type, but only a subset of \textsc{Locations} are common to both definitions).\footnote{We ignore the trivial case when an entity in $\E{A}$ is exactly identical to $\E{B}$. Without loss of generality, we also assume that entities within $\E{A}$ ($\E{B}$) are disjoint; even if they are not, we can always convert them to disjoint.}
Figure~\ref{fig:relations} illustrates these possible relationships.

\paragraph{Output:}As in \S\ref{sec:defintion}, our aim is to  train a model to recognize both $\E{A}$ and $\E{B}$. However, now a mention can get more than one entity label. Consider the \textsc{Subtype}/\textsc{Supertype} example in Figure~\ref{fig:relations}, where a mention can belong to both \{\textsc{Person}, \textsc{Actor}\} or \{\textsc{Person}, \textsc{Politician}\}, or just \textsc{Person} (i.e. \textsc{Person}, but neither \textsc{Actor} nor \textsc{Politician}). Given this, we cannot directly train an NER model over $\E{A} \cup \E{B}$. Instead in $\E{\textit{final}}$, we define an entity type for each possible combination (illustrated in the grey box in Figure~\ref{fig:relations}), and train a model over the redefined output labels. For instance, we introduce three new entity types in the final label set corresponding to the \textsc{Person} entity for the three cases discussed above--- \textsc{Person:Actor}, \textsc{Person:Politician}, and \textsc{Person:O)}. Similarly, for the \textsc{Overlapping} case, we define three new types --- \textsc{Location:A} (entities that are locations only according to $\D{A}$), \textsc{Location:B} (entities that are locations only according to $\D{B}$), and \textsc{Location:A,B} (entities that are locations according to both $\D{A}$ and $\D{B}$).

\subsection{Allowed Entity Types}
\label{subsec:allowed}
Before discussing the modifications to the methods for the non-disjoint case, we revisit the assumption that drove the methods in the disjoint case --- we observed that if a token in $\D{A}$ is annotated as O, it belongs to one of $\E{B} \cup \{\text{O}\}$, else it belongs to the annotated entity type (\S\ref{subsec:crossann}). However, this assumption does not hold in the non-disjoint setting. In Figure~\ref{fig:allowed}, \underline{\textbf{James}} is annotated as $\textsc{Person}$ in $\D{A}$, so it can belong to one of $\{$\textsc{Person:Actor}, \textsc{Person:Politician}, or \textsc{Person:O}$\}$) in the final output space.  Thus, for each token in the datasets, we define a set of \textit{allowed entity types} that the token can belong to. These allowed types are determined by the observed annotation of that token and are a combination of existing entity types or the newly-introduced types as discussed above. Figure~\ref{fig:allowed} gives more examples of allowed entity types. 

\subsection{Modifications to Proposed Methods} \label{sec:re_proposed_methods}

Given this mapping of the problem to detecting the entity types in a redefined output space, the modification to the methods from \S\ref{sec:disjoint} lies in \circled{2} from Figure~\ref{fig:methods}, where instead of annotating a token with predictions from a model, we simply constrain this annotation by the allowed entity types for that token. The rest of the steps proceed as before. We defer further details to Appendix~\ref{appendix:modifications}.
%Throughout this section, we use Fig~\ref{fig:relation_allowed_labelmaps} as our running example. Here we explain the modifications briefly, with more details in appendix~\ref{appdx:proposed_method}.

\section{Experiments}
\label{sec:experiments}

\begin{table*}[!tbp]
    \centering
        \small
\begin{tabular}{l r r r r r r} 
\toprule
\textbf{Method} & \textbf{Ontonotes} & \textbf{FewNERD-Super} & \textbf{FewNERD-Sub} & \textbf{WNUT17} &  \textbf{JNLPBA} & \textbf{I2B2}  \\
%  & & \textbf{{Super}} & \textbf{Sub} & &   &  \\
% \cline{3-5} 
% &  & \textbf{Super}& \textbf{Sub} &   &  & \\
\midrule 
% \multicolumn{8}{c}{\textbf{Baselines}} \\
% \midrule
\textbf{Naive Join} & 
76.3 \footnotesize{(2.1)} & 
68.4 \footnotesize{(2.1)} & 
54.2 \footnotesize{(1.6)} & 
27.9 \footnotesize{(3.4)} & 
54.8 \footnotesize{(3.5)} & 
76.5 \footnotesize{(2.8)}
\\ 
\textbf{CL} & 
87.7 \footnotesize{(0.2)} & 
77.7 \footnotesize{(0.1)} & 
66.1 \footnotesize{(0.2)} & 
40.6 \footnotesize{(1.3)} & 
70.5 \footnotesize{(0.5)} & 
91.2 \footnotesize{(0.3)}

\\
\textbf{AML} & 
87.4 \footnotesize{(0.3)} & 
77.9 \footnotesize{(0.1)} & 
65.6 \footnotesize{(0.2)} & 
38.4 \footnotesize{(1.7)} & 
70.6 \footnotesize{(0.4)} & 
91.6 \footnotesize{(0.3)}

\\
% \midrule
% \multicolumn{7}{c}{\textbf{Our Method}} \\
\vspace{-0.2cm}
% \midrule
\\
\textbf{X-Ann.} & 
87.8 \footnotesize{(0.2)} & 
78.3 \footnotesize{(0.1)} & 
66.9 \footnotesize{(0.1)} & 
42.3 \footnotesize{(1.2)} & 
\textbf{71.5 \footnotesize{(0.5)}} & 
92.0 \footnotesize{(0.3)}

\\
\textbf{\proposedmethod-KL} & 
\textbf{88.2 \footnotesize{(0.2)}} & 
\textbf{78.4 \footnotesize{(0.1)}} & 
67.0 \footnotesize{(0.1)} & 
\textbf{43.5 \footnotesize{(1.1)}} & 
71.4 \footnotesize{(0.5)} & 
\textbf{92.5 \footnotesize{(0.2)}}

 \\
\textbf{\proposedmethod} & 
88.1 \footnotesize{(0.2)} & 
\textbf{78.4 \footnotesize{(0.1)}} & 
\textbf{67.2 \footnotesize{(0.1)}} & 
43.3 (\footnotesize{1.3)} & 
71.4 \footnotesize{(0.5)} & 
92.3 \footnotesize{(0.3)}

\\
\vspace{-0.2cm}
\\
% \midrule
% \multicolumn{7}{c}{\textbf{Upper Bound}} \\
% \midrule
\textbf{Upper Bound} & 
88.7 \footnotesize{(0.2)} & 
78.6 \footnotesize{(0.1)} & 
67.4 \footnotesize{(0.1)} & 
45.3 \footnotesize{(0.8)} & 
71.8 \footnotesize{(0.4)} & 
93.3 \footnotesize{(0.1)} 

\\
\bottomrule
\end{tabular}
\caption{Results for the disjoint setup (mean and std. dev. micro-F1 across 25 runs). \proposedmethod~, \proposedmethodkl, and cross annotation are competitive across the board and close to the upper bound.}
\label{tab:disjoint}
\end{table*}

\subsection{Datasets, Setup, and Hyperparameters}
\label{sec:dataset_creation}
% \yvcomment{Need to add either examples or figures to explain this dataset creation process. Either in main body or appendix depending on space}

We study TE-NER using datasets covering diverse domains, entity types, and sizes---\textbf{(1)} Ontonotes \citep{weischedel2013ontonotes} \textbf{(2)} WNUT17 \citep{derczynski2017results} \textbf{(3)} JNLPBA \citep{kim2004introduction} and \textbf{(4)} I2B2  \citep{stubbs2015annotating}. Since these datasets are fully annotated, we cannot use them directly to study TE-NER. Instead we modify each dataset to obtain partial annotations. For the non-disjoint setup, we only use the FewNERD dataset. We defer more details to Appendix~\ref{appdx:dataset}.

% Different experiments require slightly different approach to create the data and the required modifications are discussed in their respective experiment sections.  

% using the a series of simple steps described below to arrive at the two partially annotated datasets $\D{A}$ and $\D{B}$, with entity types $\E{A}$ and $\E{B}$ respectively. 

% % \yogarshi{}
% \begin{enumerate}[nosep]
%     \item \textbf{Split Entity Types:} Randomly split  $\E{}$ into two (almost) equal parts and use them as $\mathcal{E_A}$ and $\mathcal{E_B}$ respectively.
%     \item \textbf{Split Dataset:} Randomly split the examples in $\mathcal{D}$ into two equal parts, call them as $\mathcal{D_A}^{\prime}$ and $\mathcal{D_B}^{\prime}$ (intermediate datasets). Note that both $\mathcal{D_A}^{\prime}$ and $\mathcal{D_B}^{\prime}$ are still fully annotated. 
%     \item \textbf{Un-Annotate $\bm{\mathcal{D_A}^{\prime}}$:} In $\mathcal{D_A}^{\prime}$, if an word is annotated as $e \in \mathcal{E_A}$, keep that annotation, otherwise remove it (make it as $\mathcal{O}$); use this modified dataset as $\mathcal{D_A}$
%     \item \textbf{Un-Annotate $\bm{\mathcal{D_B}^{\prime}}$:} Using $\mathcal{D_B}^{\prime}$ and $\mathcal{E_B}$, create $\mathcal{D_B}$ in the similar was as above
% \end{enumerate}

% \subsection{Setup and Hyperparameters}
% \label{subsec:setup}

For all experiments, we finetune BERT-base as our backbone models for all experiments with the exact setup from~\citet{devlin-etal-2019-bert}. We repeat every experiment with 5 random splits of $\D{A}$ and $\D{B}$ and 5 different seeds (for training) for each split and report mean and standard deviation of micro-F1 scores averaged across all $5\times5$ runs. For each dataset, we use the validation data to choose the best learning rate from $\{ 5e^{-6}, 1e^{-5}, 2e^{-5}, 3e^{-5}, 5e^{-5} \} $.

\subsection{Baselines and Upperbound} 

\paragraph{Naive Join:}A naive solution to TE-NER is to combine the two partially annotated datasets $\D{A}$ and $\D{B}$ and train a model on this combined dataset using a cross-entropy loss. We expect this approach to perform poorly, but it highlights the severity of issues caused by partial annotation.
\paragraph{Continual Learning (CL):}This baseline uses knowledge distillation similar to~\citet{Monaikul_Castellucci_Filice_Rokhlenko_2021} and \citet{xia-etal-2022-learn}. Briefly, we first train $\model{A}$ as teacher, then we train the student model ($\model{final}$) on $\D{B}$; if a token is annotated as $e\in\E{B}$, we use the cross-entropy loss, else we calculate KL divergence with respect to the teacher's output.   
% \yvcomment{I don't understand what the next two baselines are trying to convey. Need to follow up with Karthik.}
\paragraph{Modified Continual Learning (CL++):} Default CL does not work when the entity types are not disjoint. CL++ is a modified version of CL for our non-disjoint setup to account for allowed entity types.
%We observe that CL is just step (1), (3b) and (4) of our \proposedmethodkl method. Motivated by this, we define CL++ as (1), (3b) and (4) of our modified \proposedmethodkl.

\paragraph{Adjusted Multilabel (AML):} 
%When a token/data point can have multiple labels and/or label space keeps expanding, it is natural to pose the problem as a one-vs-all/multi-label problem over all the entity types. 
The AML baseline treats the NER problem as a multi-label classification (with sigmoid loss) problem for each token, instead of multi-class as in a standard sequence tagging approach. However, instead of all entities, the loss for multi-label considers the loss for only the allowed entity types.
% However, just like Naive Join,  naive Multilabel approach also suffers from issues with partial annotation. First, consider a case where all the entity types are disjoint; in $\mathcal{D_A}$, if a token is annotated as $O$, we know the token does not belong to $e\in \mathcal{E_A}$, but it may or may not belong to one of the entity types in $\mathcal{E_B}$, therefore, while training the model, we only use the loss terms corresponding to $e \in \mathcal{E_A}$ and mask the loss terms of $e \in \mathcal{E_B}$. Similarly for tokens annotated as $O$ in $\mathcal{D_B}$, we mask loss terms corresponding to  $\mathcal{E_A}$. The basic principle behind adjusted multilabel approach is that, if we are uncertain about a label then we do not contribute its loss term during model training. We use the same principle even when the entity types are not disjoint.

\paragraph{Upperbound:}Since $\D{A}$ and $\D{B}$ are derived from fully annotated datasets $\D{}$, where every example is annotated for $\E{\textit{final}}$, we train our model on the original, unmodified dataset $\D{}$ and use it as a hypothetical upperbound. In real scenarios, we do not have such fully annotated datasets.

\subsection{Research Questions and Experiments}
\label{subsec:rq_exp}

\begin{table*}[!tbp]
    \centering
    \small
\begin{tabular}{l r r r r  | r r r r } 

% \toprule
% \multicolumn{5}{c}{\textbf{FewNERD}} \\
\toprule
\textbf{Method} & \textbf{\textsc{Per.}} & \textbf{\textsc{Loc.}} & \textbf{\textsc{Org.}} & \textbf{\textsc{Prod.}} & \textbf{\textsc{Per.}} & \textbf{\textsc{Loc.}} & \textbf{\textsc{Org.}} & \textbf{\textsc{Prod.}} \\
\midrule 
% \multicolumn{5}{c}{\textbf{Baselines}} \\
% \midrule
\textbf{CL++} & 
73.4 \footnotesize{(0.2)} & 
76.2 \footnotesize{(0.1)} & 
75.6 \footnotesize{(0.1)} & 
77.0 \footnotesize{(0.1)} &
74.7 \footnotesize{(0.1)} & 
75.6 \footnotesize{(0.1)} &
75.1 \footnotesize{(0.1)} & 
76.8 \footnotesize{(0.1)}\\

\textbf{AML} &
72.7 \footnotesize{(0.2)} &
75.2 \footnotesize{(0.1)} &
74.5 \footnotesize{(0.1)} & 
76.5 \footnotesize{(0.1)} &
74.8 \footnotesize{(0.1)} &
75.6 \footnotesize{(0.1)} &
75.1 \footnotesize{(0.1)} & 
76.8 \footnotesize{(0.1)} \\
% \midrule
% \multicolumn{5}{c}{\textbf{Our Method}} \\
% \midrule
\vspace{-0.2cm}
\\

\textbf{X-Ann.} & 
74.1 \footnotesize{(0.1)} & 
76.9 \footnotesize{(0.1)} & 
76.3 \footnotesize{(0.1)} & 
77.7 \footnotesize{(0.1)} &
75.5 \footnotesize{(0.1)} &
\textbf{76.4 \footnotesize{(0.1)}} &
75.9 \footnotesize{(0.1)} & 
\textbf{77.6 \footnotesize{(0.1)}} \\

\textbf{\proposedmethodkl} & 
74.1 \footnotesize{(0.1)} &
76.8 \footnotesize{(0.1)} &
76.2 \footnotesize{(0.1)} &
77.7 \footnotesize{(0.1)} &
75.5 \footnotesize{(0.1)} &
76.3 \footnotesize{(0.1)} &
75.8 \footnotesize{(0.1)} & 
77.5 \footnotesize{(0.1)} \\

\textbf{\proposedmethod} & 
\textbf{74.2 \footnotesize{(0.1)}} & 
\textbf{77.0 \footnotesize{(0.1)}} &
\textbf{76.4 \footnotesize{(0.1)}}&
\textbf{77.8 \footnotesize{(0.1)}} & 
\textbf{75.6 \footnotesize{(0.2)}} &
\textbf{76.4 \footnotesize{(0.1)}} &
\textbf{76.0 \footnotesize{(0.1)}} & 
\textbf{77.6 \footnotesize{(0.1)}} \\

\vspace{-0.2cm}
% \midrule
% \multicolumn{5}{c}{\textbf{Upper Bound}}
\\
\textbf{Upper Bound} &
74.6 \footnotesize{(0.1)} &
77.1 \footnotesize{(0.1)} &
76.6 \footnotesize{(0.1)} &
78.0 \footnotesize{(0.1)} &
76.1 \footnotesize{(0.1)} &
76.9 \footnotesize{(0.1)} &
76.6 \footnotesize{(0.1)} & 
77.9 \footnotesize{(0.1)} \\
\bottomrule
\end{tabular}
\caption{Results for the \textsc{Subtype/Supertype} (left side), and \textsc{Overlapping} (right side) setups (mean and std. dev. micro-F1 across 25 runs). Each column indicates the entity type manipulated to create datasets (\S\ref{sec:dataset_creation}).}
\label{tab:supertypes_subtypes_exp}
\end{table*}

We focus on three key research questions:

\begin{enumerate}[topsep=0pt,itemsep=-1ex,partopsep=1ex,parsep=1ex]
    \item Is \proposedmethod~more accurate than other approaches given \textit{sizeable} training data ($\D{B}$) for the new entity types $\E{B}$?
    \item Is \proposedmethod~more accurate than other approaches given \textit{little} training data ($\D{B}$)?
    \item Does \proposedmethod~show robust performance when validation data is not exhaustively annotated?
\end{enumerate}

\subsubsection{Accuracy of \proposedmethod~given \textit{sizeable} $\D{B}$} \label{sec:disjoint}

We answer our first research question via experiments that focus on three different scenarios for TE-NER --- \textsc{Disjoint} (\S\ref{sec:defintion}), \textsc{Subtypes/Supertype}, and \textsc{Overlapping} (\S\ref{sec:re_defintion}). 

\paragraph{Disjoint Entity types} Table~\ref{tab:disjoint} shows the results of all methods on the disjoint setup. First, Naive Join performs significantly worse than all other methods, with a drop of as much as 16 F1 in the case of JNLPBA. This is an expected outcome, and it highlights the severity of problem caused by partial annotation. Second, both CL and AML are more accurate than the Naive Join approach as they approach the upper bound. Next, despite its simplicity, cross annotation offers a very strong solution to this problem. For 4 datasets, cross annotation is within 1 F1 of the Upper Bound and CL and AML are both behind cross annotation.

Finally, both \proposedmethod~and \proposedmethodkl~reach scores that are on par, or even slightly better than cross annotation, indicating that they offer alternative solutions to this problem. However, the differences of these methods with cross annotation are very small, and we take a closer look in future sections.

\paragraph{Non-disjont Entity Types:} We now turn to the non-disjoint setup (\S\ref{sec:revisit}), using the FewNERD dataset (\S\ref{sec:dataset_creation}). Table~\ref{tab:supertypes_subtypes_exp} shows the results of these experiments, with the left-side showing the results for the \textsc{Subtypes/Supertypes} case, and the right-side showing the results for the \textsc{Overlapping} case. In either case, each column indicates the entity type manipulated to create datasets per Appendix~\ref{appdx:dataset}. We leave out the Naive Join approach given its (expectedly) poor performance in the previous experiment.

% in which a column --- say \textsc{Person} --- represents the entity type present in $\E{A}$, whose fine-grained types are present in $\mathcal{E_B}$. We leave out the Naive Join approach given its poor performance in the previous experiment.

The observations for this set of results are very similar to the previous experiments and also consistent across the \textsc{Subtypes/Supertypes} and \textsc{Overlapping} --- cross annotation again proves to be a highly effective solution, both \proposedmethod~and \proposedmethodkl~are also on par with cross annotation, and all three methods approach the Upper Bound.

\subsubsection{Accuracy of \proposedmethod~given \textit{small} $\D{B}$}
\label{sec:exp_rq2}

\begin{table*}[!tbp]
    \small
    \centering
    \begin{tabular}{l l r r r r r r} 
        \toprule
        % \multicolumn{7}{c}{\textbf{OntoNotes}} \\
        % \midrule
        \textbf{Dataset} & \textbf{Method} & \textbf{100} & \textbf{200} & \textbf{300} & \textbf{500} & \textbf{1000} & \textbf{2000} \\
        \midrule
        \multirow{2}{*}{\textbf{OntoNotes}} &\textbf{X-Ann.} &	24.8 \footnotesize{(16.7)} & 6.2 \footnotesize{(3.1)} & 6.0 \footnotesize{(3.1)} & 57.9 \footnotesize{(6.1)} & 71.0 \footnotesize{(1.4)} & 80.1 \footnotesize{(1.0)}  \\
        &\textbf{PLM} &   37.0 \footnotesize{(24.6)} & 16.5 \footnotesize{(11.8)} & 16.5 \footnotesize{(11.6)} & 60.5 \footnotesize{(5.5)} & 75.8 \footnotesize{(1.4)} & 81.2 \footnotesize{(0.9)} \\
    \vspace{-0.2cm} \\
    \multirow{2}{*}{\textbf{FewNERD}}&  \textbf{X-Ann.} &	- & 14.4 \footnotesize{(10.0)} & 8.2 \footnotesize{(4.1)} & 42.6 \footnotesize{(7.7)} & 48.1 \footnotesize{(4.3)} & 55.9 \footnotesize{(1.4)} \\
        & \textbf{PLM} & - &28.2 \footnotesize{(4.0)} & 25.5 \footnotesize{(5.5)} & 47.4 \footnotesize{(7.8)} & 54.2 \footnotesize{(2.3)} & 60.1 \footnotesize{(0.5)} 
        \end{tabular}
    \caption{Experiments with varying sizes of $\D{B}$  (mean and std. dev. micro-F1 across 25 runs). Each column indicates the number of examples per type. As $|\D{B}|$ reduces, \proposedmethod~scores higher F1 compared to cross annotation.}
    \label{tab:effect_of_size_k_num}
\end{table*}
A reasonable assumption in TE-NER is that the dataset with the new entities ($\D{B}$) will be much smaller than the initial dataset ($\D{A}$) since it is impractical to annotate as many examples for each new entity type as exist for the old entity types. Our second set of experiments aims to test how various methods perform in scenarios where the number of examples in $\D{B}$ is limited, while maintaining the size of $\D{A}$. We only focus on cross annotation and \proposedmethod~as these were most competitive in the experiments in \S\ref{sec:disjoint}. Further, we focus only on the Ontonotes and FewNERD datasets.

% We design our experiment with varying size of $\mathcal{D_B}$ and compare the effect of size on the performance of \proposedmethod and cross annotation.  

% \noindent \textbf{Data Creation:} In step 4 --- \textbf{Un-annotate} $\bm{\mathcal{D_B^{\prime}} }$  --- instead of using all the un-annotated data, we randomly choose $K\%$ of the data and use it as $\mathcal{D_B}$. Note that size of $\mathcal{D_A}$ remains the same. Also, instead of using entire validation data, we randomly choose $K\%$ and use for validation. 

Table~\ref{tab:effect_of_size_k_num} indicates that in few-shot settings, \proposedmethod~consistently outperforms cross annotation across the board. The gap between the two methods is more severe as the number of examples reduce, with \proposedmethod~scoring as much as 10 F1 higher when $\D{B}$ contains only 100 examples for each type on Ontonotes, and 17 F1 higher on FewNERD with 300 examples per type. However, both approaches tend to be very unstable in such low data regimes, indicated by the high variance in results (more severe in Ontonotes). Despite this, \proposedmethod~exploits the soft labels from the models to achieve better scores. 

%can observe that (1) As the size of $\mathcal{D_B}$ decreases, performance of both \proposedmethod and cross annotation decreases, which is expected. (2) In low data scenario, \proposedmethod performs much better than cross-annotation; this makes sense as the \proposedmethod uses probabilities instead of hard labels like cross annotation, this is especially useful in low data scenarios as $Model_B$ might make more errors and using probabilities could mitigate the issues. Furthermore, we experimented with very less data and observed that as the data size decreases performance gap between \proposedmethod and cross annotation keeps increasing (refer Appendix~\ref{appdx:tab_effec_of_size_k_num}). 

\subsubsection{Robustness to partially annotated validation data}
\label{sec:exp_rq3}

\begin{table}[!tbp]
    \small
    \centering
    \begin{tabular}{ll rr}
        \toprule
       \textbf{Dataset} & \textbf{Val. type}& \textbf{X-Ann.} &  \textbf{\proposedmethod}\\
        \midrule
        \multirow{2}{*}{\textbf{OntoNotes}} &\textbf{Full val.} & 87.7 \footnotesize{(0.2)} & 88.1 \footnotesize{(0.2)}\\
        &\textbf{Partial val.} & 87.7 \footnotesize{(0.2)} &  88.1 \footnotesize{(0.2)}  \\
        \multirow{2}{*}{\textbf{FewNERD}} & \textbf{Full val.} & 66.8 \footnotesize{(0.2)} &  67.1 \footnotesize{(0.2)}  \\
        &\textbf{Partial val.} & 66.8 \footnotesize{(0.2)} & 66.7 \footnotesize{(0.2)} \\
        % \multicolumn{6}{c}{\textbf{OntoNotes}} \\
        % \midrule
       \bottomrule
        \end{tabular}
    \caption{F1 scores of \proposedmethod~and cross-annotation are almost identical regardless of whether validation data are partially or fully annotated.}
    \label{tab:effect_of_partial_validaton}
\end{table}

Experiments so far have assumed access to a fully annotated validation set. In practice, it is likely that even validation sets are partially annotated (similar to training sets). How do models behave with such partially annotated validation sets? Specifically, we assume that we have validation sets, $\D{A}^{val}$ and $\D{B}^{val}$ corresponding to $\D{A}$ and $\D{B}$ respectively. During the validation step, we evaluate the model being trained ($\model{final}$) on both $\D{A}^{val}$ and $\D{B}^{val}$ separately, masking out any predictions from $\E{B}$ on $\D{A}^{val}$ and from $\E{A}$ on $\D{B}^{val}$. We use the average micro-F1 on these individual validation sets as our stopping criterion. Again, we focus on cross annotation and \proposedmethod~on Ontonotes and FewNERD.
% --- lets call then $\mathcal{D_A}^{val}$ and $\mathcal{D_B}^{val}$. In this case, first  we evaluate $Model_{final}$ on $\mathcal{D_A}^{val}$, i.e. if $Model_{final}$ predicts a mention as $e \in \mathcal{E_B}$, we make it $O$ and use the modified prediction to calculate micro-F1 score on $\mathcal{D_A}^{val}$, similarly we evaluate  $Model_{final}$ on $\mathcal{D_B}^{val}$ and use the average micro-F1 score as our stopping criterion. 

Table~\ref{tab:effect_of_partial_validaton} shows that F1 scores with partial validation are similar to those of full validation. Thus, we do not even need validation data labeled with entities from both $\E{A}$ and $\E{B}$, as partially annotated validation sets are a reasonable proxy.

\subsubsection{Summary of results}

Overall, cross annotation and \proposedmethod~are similarly accurate when $\D{B}$ is large enough. Both methods are also robust when partially annotated validation sets are used instead of fully annotated validation sets. However, as the size of $\D{B}$ reduces, \proposedmethod~is increasingly more accurate.
\section{Discussion and Further Analysis} \label{sec:analysis} 

Having seen the effectiveness of cross annotation and \proposedmethod, we further analyze these methods. For these experiments, we use Ontonotes under the disjoint setting~(\S\ref{sec:disjoint}).

% In this section, we analyse the effectiveness of our \proposedmethod and cross annotation in various scenarios; for all our analysis we only consider the scenario where entity types are disjoint. 

\subsection{Effect of size of $\E{A}$ and $\E{B}$}

% \begin{table*}[!htb]
%     \centering
%     \begin{tabular}{p{3cm} c | c c c c} 
%         \toprule
%         \multicolumn{6}{c}{\textbf{OntoNotes}} \\
%         \midrule
%         \textbf{Method} & \textbf{9:9} &\textbf{12:6} & \textbf{15:3} & \textbf{16:2}& \textbf{17:1}  \\
%         \midrule
%         \textbf{Cross Annotation} &	86.52 (0.15) &	86.4 (0.19)	& 86.45 (0.27) &	86.6 (0.24)	& 86.55 (0.15)  \\
%         \textbf{\proposedmethod} & 87.01 (0.17) &	86.85 (0.19) &	86.95 (0.11) &	86.87 (0.15) &	86.85 (0.13)\\
%         \toprule
%         \multicolumn{6}{c}{\textbf{FewNERD}} \\
%         \midrule
%         & \textbf{33:33} &	\textbf{40:26} &	\textbf{50:16} &	\textbf{60:6} &	\textbf{65:1} \\
%         \midrule
%         \textbf{Cross Annotation} &	 66.86 (0.09) &	66.88 (0.17) &	66.84 (0.21) &	66.97 (0.18) &	66.85 (0.21) \\
%         \textbf{\proposedmethod} & 67.21 (0.23) &	67.1 (0.15)	& 67.11 (0.15) &	67.06 (0.11) &	67.07 (0.14)\\
%         \bottomrule
%         \end{tabular}
%     \caption{\textbf{Effect of size of $\bm{\mathcal{E_A}}$ and $\bm{\mathcal{E_B}}$: } The results are mean and std of micro-F1 score averaged across 10 runs. We can see that \proposedmethod and cross annotation are quite effective even when the size of $\mathcal{E_A}$ and $\mathcal{E_A}$ are widely different.
%     % \kkcomment{Make it a figure and change caption accordingly}
%     }
%     \label{tab:num_entity}
% \end{table*}

\begin{table}[!tbp]
\small
    \centering
    \begin{tabular}{c rr}
        \toprule
        $|\E{A}|:|\E{B}|$ & \textbf{X-Ann.} &  \textbf{\proposedmethod}\\
        \midrule
        \textbf{9:9} & 87.8 \footnotesize{(0.2)} & 88.1 \footnotesize{(0.3)}\\
        \textbf{12:6} & 87.7 \footnotesize{(0.2)} &  88.1 \footnotesize{(0.2)}  \\
        \textbf{15:3} & 87.7 \footnotesize{(0.2)} &  88.1 \footnotesize{(0.2)}  \\
        \textbf{16:2} & 87.7 \footnotesize{(0.2)} & 88.0 \footnotesize{(0.2)} \\
        \textbf{17:1} & 87.9 \footnotesize{(0.3)} & 88.1 \footnotesize{(0.2)} \\
        % \multicolumn{6}{c}{\textbf{OntoNotes}} \\
        % \midrule
       \bottomrule
        \end{tabular}
    \caption{Results with varying $|\E{A}|$ and $|\E{B}|$; F1 scores of cross annotation and \proposedmethod~do not vary with changes in $|\E{A}|$ and $|\E{B}|$.}
    \label{tab:num_entity}
\end{table}

% \begin{table*}[!htb]
%     \centering
%     \begin{tabular}{p{3cm} rrrrr} 
%         \toprule
%         % \multicolumn{6}{c}{\textbf{OntoNotes}} \\
%         % \midrule
%          & \textbf{9:9} &\textbf{12:6} & \textbf{15:3} & \textbf{16:2}& \textbf{17:1}  \\
%         \midrule
%         \textbf{Cross Annotation} &	87.8 \footnotesize{(0.2)} & 87.7 \footnotesize{(0.2)} & 87.7 \footnotesize{(0.2)} & 87.7 \footnotesize{(0.2)} & 87.9 \footnotesize{(0.3)}  \\
%         \textbf{\proposedmethod} & 88.1 \footnotesize{(0.3)} & 88.1 \footnotesize{(0.2)} & 88.1 \footnotesize{(0.2)} & 88.0 \footnotesize{(0.2)} & 88.1 \footnotesize{(0.2)} \\
%         \bottomrule
%         \end{tabular}
%     \caption{\textbf{Effect of size of $\bm{\mathcal{E_A}}$ and $\bm{\mathcal{E_B}}$: } The results are mean and std of micro-F1 score averaged across 10 runs. We can see that \proposedmethod and cross annotation are quite effective even when the size of $\mathcal{E_A}$ and $\mathcal{E_A}$ are widely different.
%     % \kkcomment{Make it a figure and change caption accordingly}
%     }
%     \label{tab:num_entity}
% \end{table*}

In experiments in \S\ref{sec:experiments}, $\E{A}$ and $\E{B}$ were set to have (almost) the same number of entity types\yvcomment{Why almost?}. However, a more likely scenario is that the number of entity types to be added ($|\E{B}|$) are fewer than the number of existing entity types ($|\E{A})|$. We investigate the behavior of methods in such settings --- for these experiments we keep the size of the datasets ($\D{A}$ and $\D{B}$) to be fixed and identical to those in \S\ref{sec:experiments}. Results in Table~\ref{tab:num_entity} indicate that even as the difference in number of entity types in $\E{A}$ and $\E{B}$ grows, cross Annotation and \proposedmethod~do not diverge in behavior and continue to be equally accurate. 

% % \noindent \textbf{Data Creation:} In step \textbf{Split Entity Type}, instead of splitting $\mathcal{E}$ --- all entity types in the dataset --- equally across $\mathcal{E_A}$ and $\mathcal{E_B}$, we (randomly) split such that $\mathcal{E_A}$ has $k$ entity types and $\mathcal{E_B}$ has rest (for multiple different $k$). Rest all the steps remain the same. Note that the size of $\mathcal{D_A}$ and $\mathcal{D_B}$ remains almost equal. 

% % Note that irrespective of how we split, the final models are comparable because (1) final entity types are same (2) they use almost the same number of annotations for each entity type; Note that the probability for an annotation to be removed --- in step $3$ and $4$ --- is still the same $50\%$

% \noindent \textbf{Result:} In table~\ref{tab:num_entity} (\kkcomment{Figure}), column $K_1\text{:}K_2$ indicate $\mathcal{E_A}$ and $\mathcal{E_B}$  has $K_1$ and $K_2$ entity types respectively. we can observe that irrespective of how we split the entity types, the final results remain almost the same. 

\subsection{Performance of $\model{\textit{final}}$ Vs $\model{B}$}

% \begin{table*}[!htb]
%     \centering
%     \begin{tabular}{p{3cm} c c c c c} 
%         \toprule
%         \multicolumn{6}{c}{\textbf{OntoNotes}} \\
%         \midrule
%         \textbf{Method} & \textbf{9:9} &\textbf{12:6} & \textbf{15:3} & \textbf{16:2}& \textbf{17:1}  \\
%         \midrule
%         \textbf{Cross Annotation} &	 -0.56 (0.44) & -0.72 (0.72) & -0.92 (1.15) & -1.71 (2.03) & -2.66 (4.66) \\
%         \textbf{\proposedmethod} & -1.15 (0.73) & -1.48 (0.81) & -2.14 (1.73) &  -2.18 (1.80) & -3.13 (4.68) \\
%         \bottomrule
%         \end{tabular}
%     \caption{\textbf{Performance of $\bm{Model_{final}}$ Vs} \textbf{ $\bm{Model_{B}}$:} The above table shows micro-F1 of $Model_{B}$ minus micro-F1 of $Model_final$ on the test set annotated with just $\mathcal{E_B}$.  We can observe that (1) $Model_{final}$ recognizes $\mathcal{E_B}$ better than $Model_B$ (2)  As the number of entity types in $\mathcal{E_A}$ increases, performance gap increases}
%     \label{tab:final_vs_B}
% \end{table*}

\begin{table}[!tbp]
    \small
    \centering
    \begin{tabular}{c rrr}
        \toprule
        $ |\E{A}|:|\E{B}| $ & $\model{B}$ & \textbf{X\-Ann.} &  
        \textbf{\proposedmethod}\\
        \midrule
        \textbf{9:9} & 87.7 \footnotesize{(2.4)} & 88.2 \footnotesize{(2.1)} & 88.5 \footnotesize{(2.1)}\\
        \textbf{12:6} &  87.3  \footnotesize{(3.1)} & 87.8 \footnotesize{(3.0)} &  88.3 \footnotesize{(2.8)}\\
        \textbf{15:3} & 84.8 \footnotesize{(8.1)} & 85.5 \footnotesize{(8.3)} & 86.3 \footnotesize{(7.1)}\\
        \textbf{16:2} & 86.0 \footnotesize{(8.9)}& 87.3 \footnotesize{(7.2)} & 87.8 \footnotesize{(6.4)}\\
        \textbf{17:1} & 72.1 \footnotesize{(13.3)} & 74.2 \footnotesize{(12.4)} & 75.7 \footnotesize{(11.9)} \\
        % \multicolumn{6}{c}{\textbf{OntoNotes}} \\
        % \midrule
       \bottomrule
        \end{tabular}
    \caption{The model from \proposedmethod~recognizes $\E{B}$ more accurately than $\model{B}$. As $|\E{A}|$ increases and $|\E{B}|$ decreases, this performance gap increases.}
    \label{tab:final_vs_B}
\end{table}

% \begin{table*}[!htb]
%     \centering
%     \begin{tabular}{p{3cm} rrrrr} 
%         \toprule
%         % \multicolumn{6}{c}{\textbf{OntoNotes}} \\
%         % \midrule
%          & \textbf{9:9} &\textbf{12:6} & \textbf{15:3} & \textbf{16:2}& \textbf{17:1}  \\
%         \midrule
%         % \multicolumn{6}{c}{\textbf{Cross Annotation}} \\
%         % \midrule
%         \textbf{Model B} & 87.7 \footnotesize{(2.4)} & 87.3  \footnotesize{(3.1)} & 84.8 \footnotesize{(8.1)} & 86.0 \footnotesize{(8.9)} & 72.1 \footnotesize{(13.3)} \\ 
%         \textbf{Cross Annotation} & 88.2 \footnotesize{(2.1)} & 87.8 \footnotesize{(3.0)} & 85.5 \footnotesize{(8.3)} & 87.3 \footnotesize{(7.2)} & 74.2 \footnotesize{(12.4)} \\
%         \textbf{\proposedmethod} & 88.5 \footnotesize{(2.1)} & 88.3 \footnotesize{(2.8)} & 86.3 \footnotesize{(7.1)} & 87.8 \footnotesize{(6.4)} & 75.7 \footnotesize{(11.9)} \\
%         \bottomrule
%         \end{tabular}
%     \caption{\textbf{Performance of $\bm{Model_{final}}$ Vs} \textbf{ $\bm{Model_{B}}$:} The above table shows micro-F1 of $Model_{B}$ minus micro-F1 of $Model_final$ on the test set annotated with just $\mathcal{E_B}$.  We can observe that (1) $Model_{final}$ recognizes $\mathcal{E_B}$ better than $Model_B$ (2)  As the number of entity types in $\mathcal{E_A}$ increases, performance gap increases}
%     \label{tab:final_vs_B}
% \end{table*}

Another possible solution to TE-NER is to use $\model{B}$ to recognize the new entity types from $\E{B}$, $\model{A}$ to recognize the original entity types $\E{A}$, and combine the predictions of these two models via heuristics. However, this begs the question --- does $\model{B}$ recognizes entities from $\E{B}$ as well as (or better than) $\model{final}$ derived from Cross Annotation or \proposedmethod? We answer this by testing these methods against $\model{B}$ on a test set annotated with just $\E{B}$. If $\model{final}$ predicts a mention as $e \in \E{A}$, we simply ignore it.  %, i.e., if a mention is annotated as $e \in \mathcal{E_B}$, we remove its annotation. Similarly, if  $Model_{final}$ predicts a mention as $e \in \mathcal{E_B}$, we make it $O$. 

% \noindent \textbf{Result:} \kkcomment{Update the results with Model A and update writing} Table~\ref{tab:final_vs_B} shows the performance of $Model_B$ minus performance of $Model_{final}$ and column $K_1\text{:}K_2$ indicates the case when $\mathcal{E_A}$ and $\mathcal{E_B}$ has $K_1$ and $K_2$ entity types respectively; the results are averaged across 10 runs. 

Results (Table~\ref{tab:final_vs_B}) indicate that \textbf{(1)} \proposedmethod~and cross annotation are consistently more accurate than $\model{B}$ at recognizing entity types from $\E{B}$, and \textbf{(2)} as $|\E{B}|$ decreases (and $|\E{A}|$ increases), the gap between all methods increases. We hypothesise that this is due to three reasons: (1) Cross Annotation and \proposedmethod~use additional data corresponding to $\E{A}$ in dataset $\D{A}$ hence their superior performance. (2) The more such additional information (about other entity types) present in $\D{A}$, the larger the performance gain. Intuitively, this implies that the ability to recognize entity types in $\E{A}$ is helping in better recognizing entities of interest in $\E{B}$.

\section{Conclusion}

We define and propose solutions for a general version of the problem of taxonomy expansion for NER. Crucially, and unlike prior work, our definition does not assume that the entity types that are being added are disjoint from existing types. We propose a novel approach, \proposedmethod, that is theoretically motivated based on a latent variable formulation of the problem, and show that prior solutions based on student-teacher settings are approximations of the loss arrived at by our method. \proposedmethod~outperforms baselines on various datasets, especially in data scarce scenarios when there is limited data available for the new entity types. In such settings, it is as much as 10-15 F1 points more accurate than the next best baseline. 

\section{Limitations}
There are many other extensions of the definition and setup for TE-NER that this work does not address. For instance, the old and the new entity types / datasets, can belong to different domains and results in such settings are likely to different than those reported in this paper, where both old and new entity types are created from the same pre-annotated dataset. Studying this, however, requires creation of appropriate datasets, which is also something that this work does not attempt to do.

For the more general definition of TE-NER that allows entity types to be related, We assume that we know these relations (i.e. they have been provided by an expert/user) and leave aside the problem of how to identify these relations in the first place. A separate body of work in taxonomy induction aims to identify such a hierarchy between entities (e.g. \cite{snow-etal-2006-semantic}).

Finally, despite its stronger theoretical foundations, the key method proposed in this work, \proposedmethod, is not more accurate than the simple cross annotation baseline in data rich scenarios, and is more suitable for data scarce scenarios. Further investigation is required to boost results for the former scenario. 

\yvcomment{Add future work and limitations}

% We also show that the approaches considered  Finally we analyse \proposedmethod and cross annotation in various scenarios. 

% The key takeaways from our experiments are (1) \proposedmethod and \proposedmethodkl consistently outperforms all other approaches, (2) Cross Annotation is simple, yet effective approach, when we have enough data, (3) Proposed methods are effective even when there is a huge disparity in sizes of entity types $A$ and $B$, (4) TE-NER might help performance, even if we are interested only in $\mathcal{E_A}$ or $\mathcal{E_B}$ (5) \proposedmethod is significantly better than cross annotation in data scarce 
% scenario (6) proposed methods perform quite well even with just partially annotated validation data.  

\bibliography{anthology,custom}

\begin{thebibliography}{28}
\expandafter\ifx\csname natexlab\endcsname\relax\def\natexlab#1{#1}\fi

\bibitem[{Arabie et~al.(1996)Arabie, Hubert, De~Soete, and
  Gordon}]{arabie1996hierarchical}
P~Arabie, L~Hubert, G~De~Soete, and A~Gordon. 1996.
\newblock Hierarchical classification.
\newblock \emph{P. Arabie, L. Hubert, G. De Soete, \& A. Gordon, Clustering and
  classification}, pages 65--121.

\bibitem[{Bekker and Davis(2020)}]{bekker2020learning}
Jessa Bekker and Jesse Davis. 2020.
\newblock Learning from positive and unlabeled data: A survey.
\newblock \emph{Machine Learning}, 109(4):719--760.

\bibitem[{Chen and Moschitti(2019)}]{chen2019transfer}
Lingzhen Chen and Alessandro Moschitti. 2019.
\newblock Transfer learning for sequence labeling using source model and target
  data.
\newblock In \emph{Proceedings of the AAAI Conference on Artificial
  Intelligence}, volume~33, pages 6260--6267.

\bibitem[{Choi et~al.(2018)Choi, Levy, Choi, and Zettlemoyer}]{choi2018ultra}
Eunsol Choi, Omer Levy, Yejin Choi, and Luke Zettlemoyer. 2018.
\newblock Ultra-fine entity typing.
\newblock In \emph{Proceedings of the 56th Annual Meeting of the Association
  for Computational Linguistics (Volume 1: Long Papers)}, pages 87--96.

\bibitem[{De~Lange et~al.(2019)De~Lange, Aljundi, Masana, Parisot, Jia,
  Leonardis, Slabaugh, and Tuytelaars}]{de2019continual}
Matthias De~Lange, Rahaf Aljundi, Marc Masana, Sarah Parisot, Xu~Jia, Ales
  Leonardis, Gregory Slabaugh, and Tinne Tuytelaars. 2019.
\newblock Continual learning: A comparative study on how to defy forgetting in
  classification tasks.
\newblock \emph{arXiv preprint arXiv:1909.08383}, 2(6):2.

\bibitem[{Derczynski et~al.(2017)Derczynski, Nichols, van Erp, and
  Limsopatham}]{derczynski2017results}
Leon Derczynski, Eric Nichols, Marieke van Erp, and Nut Limsopatham. 2017.
\newblock Results of the wnut2017 shared task on novel and emerging entity
  recognition.
\newblock In \emph{Proceedings of the 3rd Workshop on Noisy User-generated
  Text}, pages 140--147.

\bibitem[{Devlin et~al.(2019)Devlin, Chang, Lee, and
  Toutanova}]{devlin-etal-2019-bert}
Jacob Devlin, Ming-Wei Chang, Kenton Lee, and Kristina Toutanova. 2019.
\newblock \href {https://doi.org/10.18653/v1/N19-1423} {{BERT}: Pre-training of
  deep bidirectional transformers for language understanding}.
\newblock In \emph{Proceedings of the 2019 Conference of the North {A}merican
  Chapter of the Association for Computational Linguistics: Human Language
  Technologies, Volume 1 (Long and Short Papers)}, pages 4171--4186,
  Minneapolis, Minnesota. Association for Computational Linguistics.

\bibitem[{Ding et~al.(2021)Ding, Xu, Chen, Wang, Han, Xie, Zheng, and
  Liu}]{ding2021few}
Ning Ding, Guangwei Xu, Yulin Chen, Xiaobin Wang, Xu~Han, Pengjun Xie, Haitao
  Zheng, and Zhiyuan Liu. 2021.
\newblock Few-nerd: A few-shot named entity recognition dataset.
\newblock In \emph{Proceedings of the 59th Annual Meeting of the Association
  for Computational Linguistics and the 11th International Joint Conference on
  Natural Language Processing (Volume 1: Long Papers)}, pages 3198--3213.

\bibitem[{Grave(2014)}]{grave2014weakly}
Edouard Grave. 2014.
\newblock Weakly supervised named entity classification.
\newblock In \emph{Workshop on Automated Knowledge Base Construction (AKBC)}.

\bibitem[{Hinton et~al.(2015)Hinton, Vinyals, Dean
  et~al.}]{hinton2015distilling}
Geoffrey Hinton, Oriol Vinyals, Jeff Dean, et~al. 2015.
\newblock Distilling the knowledge in a neural network.
\newblock \emph{arXiv preprint arXiv:1503.02531}, 2(7).

\bibitem[{Jie et~al.(2019)Jie, Xie, Lu, Ding, and Li}]{jie-etal-2019-better}
Zhanming Jie, Pengjun Xie, Wei Lu, Ruixue Ding, and Linlin Li. 2019.
\newblock \href {https://doi.org/10.18653/v1/N19-1079} {Better modeling of
  incomplete annotations for named entity recognition}.
\newblock In \emph{Proceedings of the 2019 Conference of the North {A}merican
  Chapter of the Association for Computational Linguistics: Human Language
  Technologies, Volume 1 (Long and Short Papers)}, pages 729--734, Minneapolis,
  Minnesota. Association for Computational Linguistics.

\bibitem[{Kim et~al.(2004)Kim, Ohta, Tsuruoka, Tateisi, and
  Collier}]{kim2004introduction}
Jin-Dong Kim, Tomoko Ohta, Yoshimasa Tsuruoka, Yuka Tateisi, and Nigel Collier.
  2004.
\newblock Introduction to the bio-entity recognition task at jnlpba.
\newblock In \emph{Proceedings of the international joint workshop on natural
  language processing in biomedicine and its applications}, pages 70--75.
  Citeseer.

\bibitem[{Ling and Weld(2012)}]{ling2012fine}
Xiao Ling and Daniel~S Weld. 2012.
\newblock Fine-grained entity recognition.
\newblock In \emph{Twenty-Sixth AAAI Conference on Artificial Intelligence}.

\bibitem[{Ma et~al.(2022)Ma, Ballesteros, Doss, Anubhai, Mallya, Al-Onaizan,
  and Roth}]{ma-etal-2022-label}
Jie Ma, Miguel Ballesteros, Srikanth Doss, Rishita Anubhai, Sunil Mallya, Yaser
  Al-Onaizan, and Dan Roth. 2022.
\newblock \href {https://doi.org/10.18653/v1/2022.findings-acl.155} {Label
  semantics for few shot named entity recognition}.
\newblock In \emph{Findings of the Association for Computational Linguistics:
  ACL 2022}, pages 1956--1971, Dublin, Ireland. Association for Computational
  Linguistics.

\bibitem[{Mai et~al.(2018)Mai, Pham, Nguyen, Nguyen, Bollegala, Sasano, and
  Sekine}]{mai-etal-2018-empirical}
Khai Mai, Thai-Hoang Pham, Minh~Trung Nguyen, Tuan~Duc Nguyen, Danushka
  Bollegala, Ryohei Sasano, and Satoshi Sekine. 2018.
\newblock \href {https://aclanthology.org/C18-1060} {An empirical study on
  fine-grained named entity recognition}.
\newblock In \emph{Proceedings of the 27th International Conference on
  Computational Linguistics}, pages 711--722, Santa Fe, New Mexico, USA.
  Association for Computational Linguistics.

\bibitem[{Mayhew et~al.(2019)Mayhew, Chaturvedi, Tsai, and
  Roth}]{mayhew-etal-2019-named}
Stephen Mayhew, Snigdha Chaturvedi, Chen-Tse Tsai, and Dan Roth. 2019.
\newblock \href {https://doi.org/10.18653/v1/K19-1060} {Named entity
  recognition with partially annotated training data}.
\newblock In \emph{Proceedings of the 23rd Conference on Computational Natural
  Language Learning (CoNLL)}, pages 645--655, Hong Kong, China. Association for
  Computational Linguistics.

\bibitem[{Meng et~al.(2019)Meng, Shen, Zhang, and Han}]{meng2019weakly}
Yu~Meng, Jiaming Shen, Chao Zhang, and Jiawei Han. 2019.
\newblock Weakly-supervised hierarchical text classification.
\newblock In \emph{Proceedings of the AAAI conference on artificial
  intelligence}, volume~33, pages 6826--6833.

\bibitem[{Monaikul et~al.(2021)Monaikul, Castellucci, Filice, and
  Rokhlenko}]{Monaikul_Castellucci_Filice_Rokhlenko_2021}
Natawut Monaikul, Giuseppe Castellucci, Simone Filice, and Oleg Rokhlenko.
  2021.
\newblock \href {https://ojs.aaai.org/index.php/AAAI/article/view/17600}
  {Continual learning for named entity recognition}.
\newblock \emph{Proceedings of the AAAI Conference on Artificial Intelligence},
  35(15):13570--13577.

\bibitem[{Peng et~al.(2019)Peng, Xing, Zhang, Fu, and
  Huang}]{peng-etal-2019-distantly}
Minlong Peng, Xiaoyu Xing, Qi~Zhang, Jinlan Fu, and Xuanjing Huang. 2019.
\newblock \href {https://doi.org/10.18653/v1/P19-1231} {Distantly supervised
  named entity recognition using positive-unlabeled learning}.
\newblock In \emph{Proceedings of the 57th Annual Meeting of the Association
  for Computational Linguistics}, pages 2409--2419, Florence, Italy.
  Association for Computational Linguistics.

\bibitem[{Phang et~al.(2018)Phang, F{\'e}vry, and Bowman}]{Phang2018SentenceEO}
Jason Phang, Thibault F{\'e}vry, and Samuel~R. Bowman. 2018.
\newblock Sentence encoders on stilts: Supplementary training on intermediate
  labeled-data tasks.
\newblock \emph{ArXiv}, abs/1811.01088.

\bibitem[{Ramshaw and Marcus(1995)}]{ramshaw-marcus-1995-text}
Lance Ramshaw and Mitch Marcus. 1995.
\newblock \href {https://aclanthology.org/W95-0107} {Text chunking using
  transformation-based learning}.
\newblock In \emph{Third Workshop on Very Large Corpora}.

\bibitem[{Ringland et~al.(2019)Ringland, Dai, Hachey, Karimi, Paris, and
  Curran}]{ringland2019nne}
Nicky Ringland, Xiang Dai, Ben Hachey, Sarvnaz Karimi, Cecile Paris, and
  James~R Curran. 2019.
\newblock Nne: A dataset for nested named entity recognition in english
  newswire.
\newblock In \emph{Proceedings of the 57th Annual Meeting of the Association
  for Computational Linguistics}, pages 5176--5181.

\bibitem[{Shmelkov et~al.(2017)Shmelkov, Schmid, and
  Alahari}]{shmelkov2017incremental}
Konstantin Shmelkov, Cordelia Schmid, and Karteek Alahari. 2017.
\newblock Incremental learning of object detectors without catastrophic
  forgetting.
\newblock In \emph{Proceedings of the IEEE international conference on computer
  vision}, pages 3400--3409.

\bibitem[{Silla and Freitas(2011)}]{silla2011survey}
Carlos~N Silla and Alex~A Freitas. 2011.
\newblock A survey of hierarchical classification across different application
  domains.
\newblock \emph{Data Mining and Knowledge Discovery}, 22(1):31--72.

\bibitem[{Snow et~al.(2006)Snow, Jurafsky, and Ng}]{snow-etal-2006-semantic}
Rion Snow, Daniel Jurafsky, and Andrew~Y. Ng. 2006.
\newblock \href {https://doi.org/10.3115/1220175.1220276} {Semantic taxonomy
  induction from heterogenous evidence}.
\newblock In \emph{Proceedings of the 21st International Conference on
  Computational Linguistics and 44th Annual Meeting of the Association for
  Computational Linguistics}, pages 801--808, Sydney, Australia. Association
  for Computational Linguistics.

\bibitem[{Stubbs and Uzuner(2015)}]{stubbs2015annotating}
Amber Stubbs and {\"O}zlem Uzuner. 2015.
\newblock Annotating longitudinal clinical narratives for de-identification:
  The 2014 i2b2/uthealth corpus.
\newblock \emph{Journal of biomedical informatics}, 58:S20--S29.

\bibitem[{Weischedel et~al.(2013)Weischedel, Palmer, Marcus, Hovy, Pradhan,
  Ramshaw, Xue, Taylor, Kaufman, Franchini et~al.}]{weischedel2013ontonotes}
Ralph Weischedel, Martha Palmer, Mitchell Marcus, Eduard Hovy, Sameer Pradhan,
  Lance Ramshaw, Nianwen Xue, Ann Taylor, Jeff Kaufman, Michelle Franchini,
  et~al. 2013.
\newblock Ontonotes release 5.0 ldc2013t19.
\newblock \emph{Linguistic Data Consortium, Philadelphia, PA}, 23.

\bibitem[{Xia et~al.(2022)Xia, Wang, Lyu, Zhu, Wu, Li, and
  Dai}]{xia-etal-2022-learn}
Yu~Xia, Quan Wang, Yajuan Lyu, Yong Zhu, Wenhao Wu, Sujian Li, and Dai Dai.
  2022.
\newblock \href {https://doi.org/10.18653/v1/2022.findings-acl.179} {Learn and
  review: Enhancing continual named entity recognition via reviewing synthetic
  samples}.
\newblock In \emph{Findings of the Association for Computational Linguistics:
  ACL 2022}, pages 2291--2300, Dublin, Ireland. Association for Computational
  Linguistics.

\end{thebibliography}

\appendix
% \section{Taxonomy Expansion: Formal Definition} \label{appdx:defintion}

\section{Dataset creation process}
\label{appdx:dataset}

We create partially annotated datasets from a fully annotated dataset $\D{}$ with entity types $\E{}$ by first splitting the $\E{}$ into two (approximately) equal subsets (yielding $\E{A}$ and $\E{B}$), and then repeating this with the examples in $\D{}$ (yielding $\D{A}$ and $\D{B}$). Additionally, for the examples, we also scrub the annotations corresponding to the complementary entity types (e.g., remove entity annotations corresponding to $\E{A}$ from $\D{B}$). 

For validation and test data, we use the corresponding fully annotated development and test sets as it is respectively. However, the assumption of fully annotated validation data being available is unrealistic, and we return to this in $\S\ref{sec:exp_rq3}$

\paragraph{Datasets for Non-disjoint TE-NER } 

We study the non-disjoint version of the problem (\S\ref{sec:re_defintion}) using the FewNERD dataset~\citep{ding2021few}. This dataset consists of two levels of annotations corresponding to coarse-grained supertypes (e.g., \textsc{Person}) and fine grained subtypes (e.g., \textsc{Actor}).\footnote{For the disjoint version of the problem (\S\ref{sec:disjoint}), we simply treat these two different levels of the dataset as two different datasets ---  FewNERD-Super and FewNERD-Sub.} 

To study the \textsc{Subtype}/\textsc{Supertype} scenario, after we split the coarse-grained entity types $\E{}$ into $\E{A}$ and $\E{B}$ as above, we randomly add a subset of the subtypes corresponding to a coarse-grained entity type $e \in \E{A}$ to $\E{B}$. For instance, if the \textsc{Person} entity type is present in $\E{A}$, we add a subset of its eight subtypes (\textsc{Politician}, \textsc{Actor}, \textsc{Artist}) etc.) to $\E{B}$.

To study the \textsc{Overlapping} scenario, we first choose a coarse-grained entity type (e.g. \textsc{Person}), split its fine-grained subtypes into two overlapping subsets $S_A$ and $S_B$ (e.g. $S_A$ = {\textsc{Actor}, \textsc{Politician}, ...}, and $S_B$ = {\textsc{Actor}, \textsc{Artist}, ...}). Then, if a mention in $\D{A}$ is annotated with a subtype in $S_A$, we assign it a new entity type $\textsc{Person}_A$. Similarly, if a mention in $\D{B}$ is annotated with a subtype in $S_B$, we assign it a new entity type $\textsc{Person}_B$. $\textsc{Person}_B$ and $\textsc{Person}_B$ are added to $\E{A}$ and $\E{B}$ respectively.  

\section{Proof of lower bound} \label{appdx:proposed_method}

To show that the loss term for \proposedmethod~in Equation~\ref{eq:keyloss} is a lower bound on the KL-divergence loss used in prior work (Equation~\ref{eq:lowerbound}), we start with the loss in Equation~\ref{eq:keyloss}.
\begin{align}
    Loss(s|\theta) &=  \sum_{i:y_i = O} CE \Big(f_i(s|\theta),\ y_i \Big) \nonumber \\
    &+ \sum_{i:y_i \neq O} -\log{ \Big \langle \ f_i(s|\theta),\ g_i(s)\ \Big\rangle }.
\end{align}
Using Jensen's inequality for concave functions (like $\log{x}$)), we get $\log{E[x]} \geq E[\log{x}]$. Hence,
\begin{align}
    Loss(s|\theta) &\leq  \sum_{i:y_i = O} CE \Big(f_i(s|\theta),\ y_i \Big) \nonumber \\
    &+ \sum_{i:y_i \neq O} { \Big \langle -\log (f_i(s|\theta)),\ g_i(s)\ \Big\rangle }.\nonumber \\
    &\leq  \sum_{i:y_i = O} CE \Big(f_i(s|\theta),\ y_i \Big) \nonumber \\
    &+ \sum_{i:y_i \neq O} { \Big \langle -\log (\frac{f_i(s|\theta)}{g_i(s)}),\ g_i(s)\ \Big\rangle }.  \nonumber \\
     & =  \sum_{i:y_i = O} CE \Big(f_i(s|\theta),\ y_i \Big) \nonumber \\
    &+ \sum_{i:y_i \neq O}  KL\Big ( g_i(s) \ || \ f_i(s|\theta) \Big). 
\end{align}
% \begin{align*}
%     L^{PLM}(x, y| \Theta_{final}, g) &= \sum_{i=1}^l -\log{\langle f_i(x, \Theta_{final}), \; g_i(x, y) \rangle} \\
%     \begin{split}
%         &= \sum_{i: e_i \neq O} -\log{\langle f_i(x, \Theta_{final}), \; g_i(x, y) \rangle} \\
%         & \quad + \sum_{i: e_i = O} -\log{\langle f_i(x, \Theta_{final}), \; g_i(x, y) \rangle} \\
%     \end{split}\\
%     \begin{split}
%         &= \sum_{i: e_i \neq O} CE\Big( f_i(x, \Theta_{final}), \; e_i \Big) \\
%         & \quad + \sum_{i: e_i = O} -\log{\langle f_i(x, \Theta_{final}), \; g_i(x, y) \rangle} \\
%     \end{split}
% \end{align*}

% Now lets apply, Jensen’s Inequality: 

% \kkcomment{Complete the proof of lowerbound with KL loss function}

\section{Modifications to proposed methods}
\label{appendix:modifications}

\subsection{Modified Cross Annotation}

Recall that the idea behind cross-annotation is to use model predictions as a proxy for actual annotations. For the non-disjoint case, if a token in $\D{A}$ is annotated as $e$, then we first get its allowed entity types following \S\ref{subsec:allowed}. If this set of allowed entities contains just one element, then we annotate it as the allowed entity type. Otherwise, we use $\model{B}$ to chose the best label among the allowed entities. For the example in Figure~\ref{fig:allowed}, \textbf{\underline{James}} is annotated as \textsc{Person} in $\D{A}$, so we know that its allowed entity types are $\{$\textsc{Person:Actor}, \textsc{Person:Politician}, \textsc{Person:O} $\}$. We use $\model{B}$ to get the probability of \textsc{Actor}, \textsc{Politician} and $O$ and among them, choose the one with highest probability. After re-annotating both $\D{A}$ and $\D{B}$, the rest of the steps proceed as in \S\ref{subsec:crossann}.

\subsection{Modified PLM}

For \proposedmethod, if we are given the distribution of true labels ($g_i(s)$), the likelihood calculation does not change, therefore the loss in Equation~\ref{eq:loss_dotproduct} remains the same for the non-disjoint case. However, similar to cross annotation above, the estimation of the oracle function $g(.)$ changes. For a data point in $\D{A}$ (similarly for $\D{B}$), if a token is annotated as $e$, we need $g(.)$ to give a probability distribution over its allowed entity types (the probability of other entity types is 0). 
%However, we can not directly use the corresponding probabilities from $\model{B}$ as they may not sum up to $1$. 
Therefore, we compute a softmax over the logits of the allowed entity types. In Figure~\ref{fig:allowed}, \textbf{\underline{James}} is annotated as \textsc{Person}, so we use $\model{B}$ to get the logits of \textsc{Actor}, \textsc{Politician}, and O, and then take softmax over just these (the probability of other entities in $\E{final}$ is 0)  

\end{document}